\documentclass[11pt,a4paper,final]{article}
\usepackage{fontenc}
\usepackage[utf8]{inputenc}
\usepackage{textcomp}
\usepackage[usenames,dvipsnames]{color}
\usepackage{url}
\usepackage[english]{babel}
\usepackage[final]{graphicx}
\usepackage{showkeys}
\usepackage[font=scriptsize, labelfont=bf]{caption}
\usepackage{fancyhdr}
\providecommand{\keywords}[1]{\textbf{keywords} #1}

\usepackage{amsfonts, amssymb, mathrsfs, amsmath}
\usepackage[draft]{fixme}

\title{Most Likely Separation of Intensity and Warping Effects in Image Registration\thanks{This work was
supported by the CSGB Centre for Stochastic
Geometry and Advanced Bioimaging funded by a grant from the Villum foundation.}} 

\author{Line K\"uhnel{\footnotemark[2]}
, Stefan Sommer{\footnotemark[2]}
, Akshay Pai{\thanks{Department of Computer Science, University of Copenhagen, Denmark (kuhnel@di.ku.dk/ sommer@di.ku.dk/ akshay@di.ku.dk)}} and Lars Lau Raket{\thanks{Department of Mathematical Sciences, University of Copenhagen, Denmark (larslau@math.ku.dk)}}}

\newcommand{\ud}{\mathrm{d}} 
\newcommand{\RR}{\mathbb{R}} 

\newcommand{\II}{\mathbb{I}} 

\newcommand{\by}{\boldsymbol{y}}

\newcommand{\btheta}{\boldsymbol{\theta}}

\newcommand{\bx}{{\boldsymbol{x}}}

\newcommand{\bw}{\boldsymbol{w}}

\usepackage{todonotes}

\usepackage[ruled]{algorithm2e}
\usepackage{color}

\begin{document}

\maketitle

\begin{abstract}
This paper introduces a class of mixed-effects models for joint modeling of spatially correlated intensity variation and warping variation in 2D images. Spatially correlated intensity variation and warp variation are modeled as random effects, resulting in a nonlinear mixed-effects model that enables simultaneous estimation of template and model parameters by optimization of the likelihood function.
We propose an algorithm for fitting the model which alternates estimation of variance parameters and image registration. This approach avoids the potential estimation bias in the template estimate that arises when treating registration as a preprocessing step.
 We apply the model to datasets of facial images and 2D brain magnetic resonance images to illustrate the simultaneous estimation and prediction of intensity and warp effects.
\end{abstract}



\pagestyle{myheadings}
\thispagestyle{plain}
\markboth{K\"UHNEL, SOMMER, PAI \& RAKET}{SEPARATING INTENSITY AND WARPING EFFECTS}

\section{Introduction}
When analyzing collections of imaging data, a general goal is to quantify similarities and differences across images. In medical image analysis and computational anatomy, a common goal is to find patterns that can distinguish morphologies of healthy and diseased subjects aiding the understanding of the population epidemiology. Such distinguishing patterns are typically investigated by comparing single observations to a representative member of the underlying population, and statistical analyses are performed relative to this representation. In the context of medical imaging, it has been customary to choose the template from the observed data as a common image of the population. However, such an approach has been shown to be highly dependent on the choice of the image. In more recent approaches, the templates are estimated using statistical methods that make use of the additional information provided by the observed data~\cite{ma_bayesian_2008}.

In order to quantify the differences between images, the dominant modes of variation in the data must be identified. Two major types of variability in a collection of comparable images are \emph{intensity variation} and variation in \emph{point-correspondences}. Point-correspondence or \emph{warp} variation can be viewed as shape variability of an individual observation with respect to the template. Intensity variation is the variation that is left when the observations are compensated for the true warp variation. This typically includes noise artifacts like systematic error and sensor noise or anatomical variation such as tissue density or tissue texture. Typically one would assume that the intensity variation consists of both independent noise and spatially correlated effects.

\begin{figure}[!th]
\centering
\mbox{
 \hspace{1.8cm}
\def\svgwidth{\columnwidth}
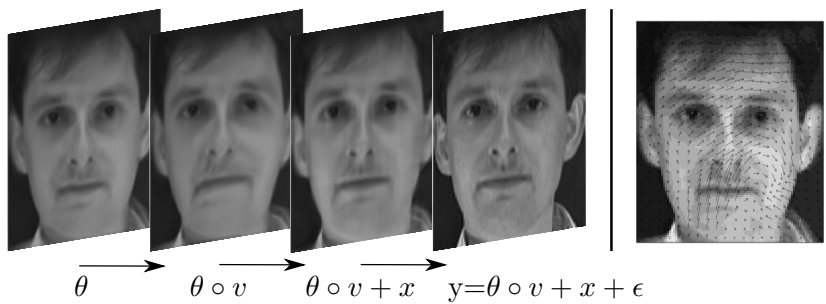
}
\caption{Fixed and random effects:
    The template ($\theta$: leftmost) pertubed by 
    random warp ($\theta\circ v$: 2nd from left) and warp+spatially correlated intensity ($\theta\circ
    v+x$: 3rd from left) together with independent noise $\epsilon$ constitute the
    observation ($y$: 4th from left). Right: the warp field $v$
  that brings the observation into spatial correspondence with $\theta$ overlayed the template. 
  Estimation of template and model hyperparameters are
  conducted simultaneously with prediction of random effects allowing separation
of the different factors in the nonlinear model.}\label{fig:layer}
\end{figure}

In this work, we introduce a flexible class of mixed-effects models that explicitly model the template as a fixed effect and intensity and warping variation as random effects, see Figure~\ref{fig:layer}. This simultaneous approach enables separation of the random variation effects in a data-driven fashion using alternating maximum-likelihood estimation and prediction. The resulting model will therefore choose the separation of intensity and warping effects that is most likely given the patterns of variation found in the data. From the model specification and estimates, we are able to denoise observations through linear prediction in the model under the maximum likelihood estimates. Estimation in the model is performed with successive linearization around the warp parameters enabling the use of linear mixed-effects predictors and avoiding the use of sampling techniques to account for nonlinear terms. We apply our method on datasets of face images and 2D brain MRIs to illustrate its ability to estimate templates for populations and predict warp and intensity effects.

\subsection{Outline of the paper}
The paper is structured as follows. In Section~\ref{sec:background}, we give an overview of previously introduced methods for analyzing image data with warp variation. Section~\ref{sec:mod} covers the mixed-effects model including a description of the estimation procedure (Section~\ref{sec:est}) and how to predict from the model (Section~\ref{sec:pred}). In Section~\ref{sec:invB}, we give an example of how to model spatially correlated variations with a tied-down Brownian sheet. We consider two applications of the mixed-effects model to real-life datasets in Section~\ref{sec:app} and Section~\ref{SimStud} contains a simulation study that is used for comparing the precision of the model to more conventional approaches. 

\section{Background}
\label{sec:background}
The model introduced in this paper focuses on separately modelling the intensity and warp variation. Image registration conventionally only focuses on identifying warp differences between pairs of images. The intensity variation is not included in the model and possible removal of this effect is considered as a pre-or postprocessing step. The warp differences are often found by solving a variational problem of the form
\begin{equation}
  E_{I_1,I_2}(\varphi)
  =
  R(\varphi)
  +
  \lambda S(I_1,I_2\circ\varphi^{-1})
  ,\label{eq:varprob}
\end{equation}
see for example \cite{sotiras_deformable_2013}.
Here $S$ measures the dissimilarity between the fixed image $I_1$ and the warped image $I_2\circ\varphi^{-1}$, $R$ is a regularization on the warp $\varphi$, and $\lambda>0$ is a weight that is often chosen by ad-hoc methods. After registration, either the warp, captured in $\varphi$, or the intensity differences between $I_1$ and $I_2\circ\varphi^{-1}$ can be analyzed~\cite{su_cortical_2016}. Several works have defined methods that incorporate registration as part of the defined models. The approach described in this paper will also regard registration as a part of the proposed model and adress the following three problems that arise in image analysis: (a) being able to estimate model parameters such as $\lambda$ in a data-driven fashion;
(b) assuming a generative statistical model that gives explicit interpretation of the terms that corresponds to the dissimilarity $S$ and penalization $R$; and (c) being simultaneous in the estimation of population-wide effects such as the mean or template image and individual per-image effects, such as the warp and intensity effects. These features are of fundamental importance in image registration and many works have addressed combinations of them. The main difference of our approach to state-of-the-art statistical registration frameworks is that we propose a simultaneous random model for warp and intensity variation. As we will see, the combination of maximum likelihood estimation and the simultaneous random model for warp and intensity variation manifests itself in a trade-off where the uncertainty of both effects are taken into account simultaneously. As a result, when estimating fixed effects and predicting random effects in the model the most likely separation of the effects given the observed patterns of variation in the entire data material is used.

Methods for analyzing collections of image data, for example template estimation in medical imaging \cite{joshi_unbiased_2004}, with both intensity and warping effects can be divided into two categories, \emph{two-step methods} and \emph{simultaneous methods}. Two-step methods perform alignment as a preprocessing step before analyzing the aligned data. Such methods can be problematic because the data is modified and the uncertainty related to the modification is ignored in the subsequent analysis. This means that the effect of intensity variation is generally underestimated, which can introduce bias in the analysis, see \cite{RaketSommerMarkussen} for the corresponding situation in 1D functional data analysis. Simultaneous methods, on the other hand, seek to analyze the images in a single step that includes the alignment procedure. 

Conventional simultaneous methods typically use $L^2$ data terms to measure dissimilarity. Such dissimilarity measures are equivalent to the model assumption that the intensity variation in the image data consists solely of uncorrelated Gaussian noise. This approach is commonly used in image registration with the sum of squared differences (SSD) dissimilarity measure, and in atlas estimation \cite{zhang_bayesian_2013}. Since the $L^2$ data term is very fragile to systematic deviations from the model assumption, for example contrast differences, the method can perform poorly. One solution to make the $L^2$ data term more robust against systematic intensity variation and in general to insufficient information in the data term is to add a strong penalty on the variation of the warping functions. This approach is however an implicit solution to the problem, since the gained robustness is a side effect of regularizing another model component. As a consequence, the effect on the estimates is very hard to quantify, and it is very hard to specify a suitable regularization for a specific type of intensity variation. This approach is, for example, taken in the variational formulations of the template estimation problem in \cite{joshi_unbiased_2004}. An elegant instance of this strategy is the Bayesian model presented in \cite{allassonniere2015bayesian} where the warping functions are modeled as latent Gaussian effects with an unknown covariance that is estimated in a data-driven fashion. Conversely, systematic intensity variation can be sought to be removed prior to the analysis, in a reversed two-step method, for example by using bias-correction techniques for MRI data \cite{tustison_n4itk:_2010}. The presence of warp variation can however influence the estimation of the intensity effects.

Analysis of images with systematic intensity differences can be improved using data dissimilarity measures that are robust or invariant to such systematic differences. However, robustness and invariance come at a cost in accuracy. By choosing a specific kind of invariance in the dissimilarity measure, the model is given a pre-specified recipe for separating intensity and warping effects; the warps should maximize the invariant part of the residual under the given model parameters. Examples of classical robust data terms include $L^1$-norm data terms \cite{pock2007duality}, Charbonnier data terms \cite{bruhn2005lucas}, and Lorentzian data terms \cite{blackanandan}. Robust data terms are often challenging to use, since they may not be differentiable ($L^1$-norms) or may not be convex (Lorentzian data term). A wide variety of invariant data terms have been proposed, and are useful when the invariances represent a dominant mode of variation in the data. Examples of classical data terms that are invariant to various linear and nonlinear photometric relationships are normalized cross-correlation, correlation-ratio and mutual information \cite{maes1997multimodality, hermosillo2002variational, roche1998correlation, panin2012mutual}. Another approach for achieving robust or invariant data terms is to transform the data that is used in the data term. A classical idea is to match discretely computed gradients or other discretized derivative quantities \cite{Papenberg}. A related idea is to construct invariant data terms based on discrete transformations. This type of approach has become increasingly popular in image matching in recent years. Examples include the rank transform and the census transform \cite{zabih1994non, mohamed2012tv, hafner2013census, hafner2015mathematical}, and more recently the complete rank transform \cite{demetz2013complete}. While both robust and invariant data terms have been shown to give very good results in a wide array of applications, they induce a fixed measure of variation that does not directly model variation in the data. Thus, the general applicability of the method can come at the price of limited accuracy.

Several alternative approaches for analyzing warp and intensity simultaneously have been proposed \cite{negahdaripour_revised_1998, jorstad_deformation_2011, blanz_morphable_1999, xie_face_2008}. In \cite{negahdaripour_revised_1998} warps between images are considered as combination of two transformation fields, one representing the image motion (warp effect) and one describing the change of image brightness (intensity effect). Based on this definition warp and intensity variation can be modeled simultaneously.  An alternative approach is considered in \cite{jorstad_deformation_2011}, where an invariant metric is used, which enables analysis of the dissimilarity in point correspondences between images disregarding the intensity variation. These methods are not statistical in the sense that they do not seek to model the random structures of the variation of the image data. A statistical model is presented in \cite{blanz_morphable_1999}, where parameters for texture, shape variation (warp) and rendering are estimated using maximizing-a-posteriori estimation. 

To overcome the mentioned limitations of conventional approaches, we propose to do statistical modeling of the sources of variation in data.
By using a statistical model where we assume parametric covariance structures for the different types of observed variation, the variance parameters can be estimated from the data. The contribution of different types of variation is thus weighted differently in the data term. By using, for example, maximum-likelihood estimation, the most likely form of the variation given the data is penalized the least. We emphasize that in contrast to previous mixed-effects models incorporating warp effects \cite{allassonniere2015bayesian,zhang_bayesian_2013}, the goal here is to simultaneously model warp and intensity effects. These effects impose randomness relative to a template, the fixed-effect, that is estimated during the inference process.

The nonlinear mixed-effects models are a commonly used tool in statistics. These types of models can be computationally intensive to fit, and are rarely used for analyzing large data sizes such as image data. We formulate the proposed model as a nonlinear mixed-effects model and demonstrate how certain model choices can be used to make estimation in the model computationally feasible for large data sizes. The model incorporates random intensity and warping effects in a small-deformation setting: We do not require warping functions to produce diffeomorphisms. The geometric structure is therefore more straightforward  than in for example the LDDMM model \cite{younes2010shapes}. From a statistical perspective, the small-deformation setting is much easier to handle than the large-deformation setting where warping functions are restricted to produce diffeomorphisms.  

 Instead of requiring diffeomorphisms, we propose a class of models that will produce warping functions that in most cases do not fold. Another advantage of the small-deformation setting is that we can model the warping effects as latent Gaussian disparity vectors in the domain. Such direct modeling allows one to compute a high-quality approximation of the likelihood function by linearizing the model around the modes of the nonlinear latent random variables. The linearized model can be handled using conventional methods for linear mixed-effects models \cite{pinheiro2006mixed} which are very efficient compared to sampling-based estimation procedures. 

In the large-deformation setting, the metamorphosis model \cite{trouve_local_2005,trouve_metamorphoses_2005} extends the LDDMM framework for image registration \cite{younes2010shapes} to include intensity change in images. Warp and intensity differences are modeled separately in metamorphosis with a Riemannian structure measuring infinitesimal variation in both warp and intensity. While this separation has similarities to the statistical model presented here, we are not aware of any work which have considered likelihood-based estimation of variables in metamorphosis models.

\section{Statistical model}
\label{sec:mod}
We consider spatial functional data defined on $\RR^2$ taking values in $\RR$. Let $\boldsymbol{y}_1,\ldots,\boldsymbol{y}_n$ be $n$ functional observations on a regular lattice with $m = m_1m_2$ points $(s_j,t_k)$, that is,  $\boldsymbol{y}_i=(y_i(s_j,t_k))_{j,k}$ for $j=1,\ldots,m_1$, $k=1,\ldots,m_2$. Consider the model in the image space
\begin{align}
y_i(s_j,t_k) = \theta(v_i(s_j,t_k)) + x_i(s_j,t_k) + \varepsilon_{ijk},
\label{eq:model}
\end{align}
for $i=1,\ldots,n$, $j=1,\ldots,m_1$ and $k=1,\ldots,m_2$. Here $\theta\colon\RR^2\to\RR$ denotes the template and $v_i\colon\RR^2\to\RR^2$ is a warping function matching a point in $y$ to a point in the template $\theta$. Moreover $x_i$ is the random spatially correlated intensity variation for which we assume that $\boldsymbol{x}_i = (x_i(s_j,t_k))_{j,k}\sim\mathcal{N}(0,\sigma^2S)$ where the spatial correlation is determined by the covariance matrix $S$. The term $\varepsilon_{ijk}\sim\mathcal{N}(0,\sigma^2)$ models independent noise. The template $\theta$ is a fixed-effect while $v_i$, $x_i$, and $\varepsilon_{ijk}$ are random.

\begin{figure}[!th]
\centering
\includegraphics[scale = 0.3]{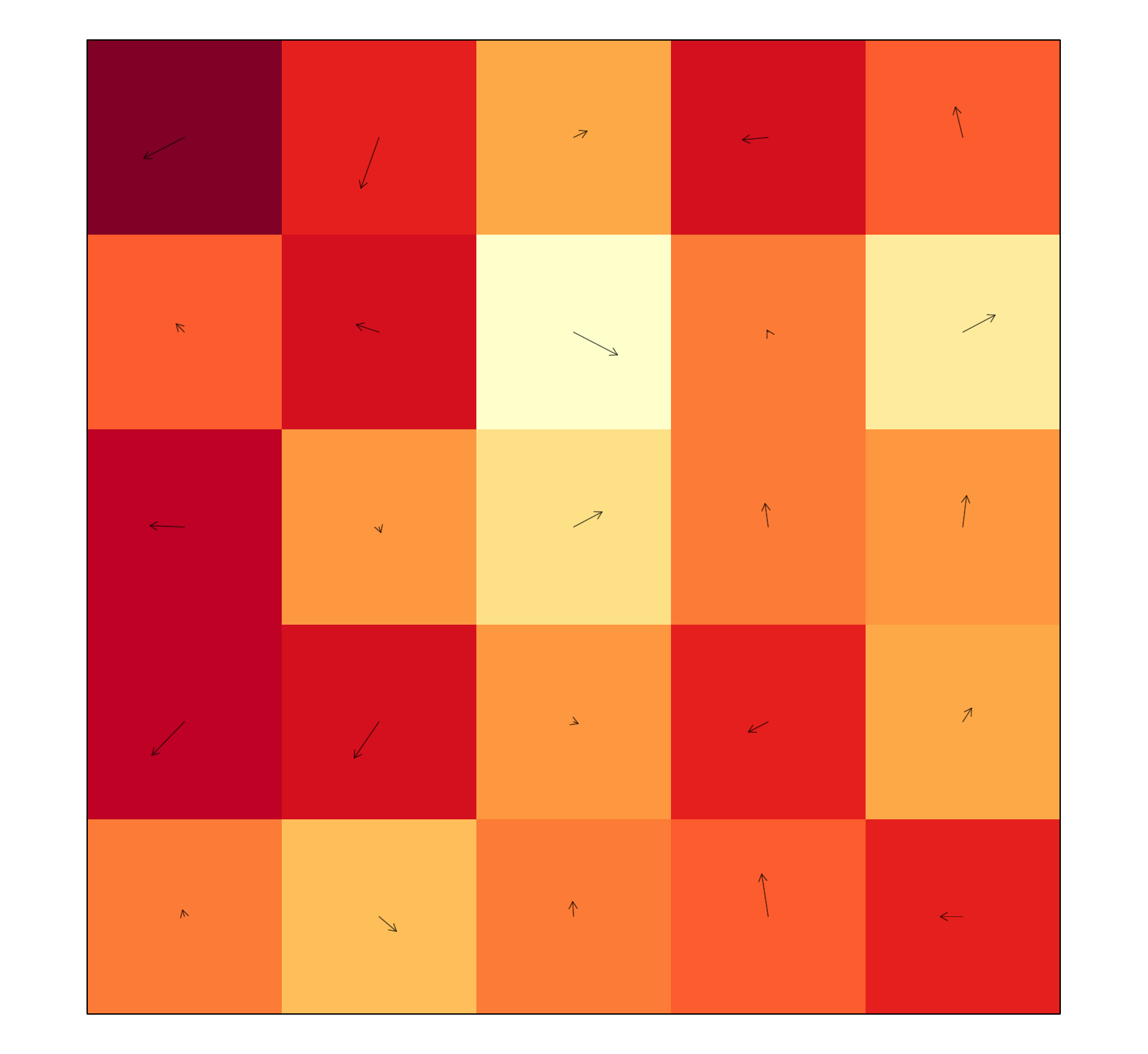} 
\includegraphics[scale = 0.3]{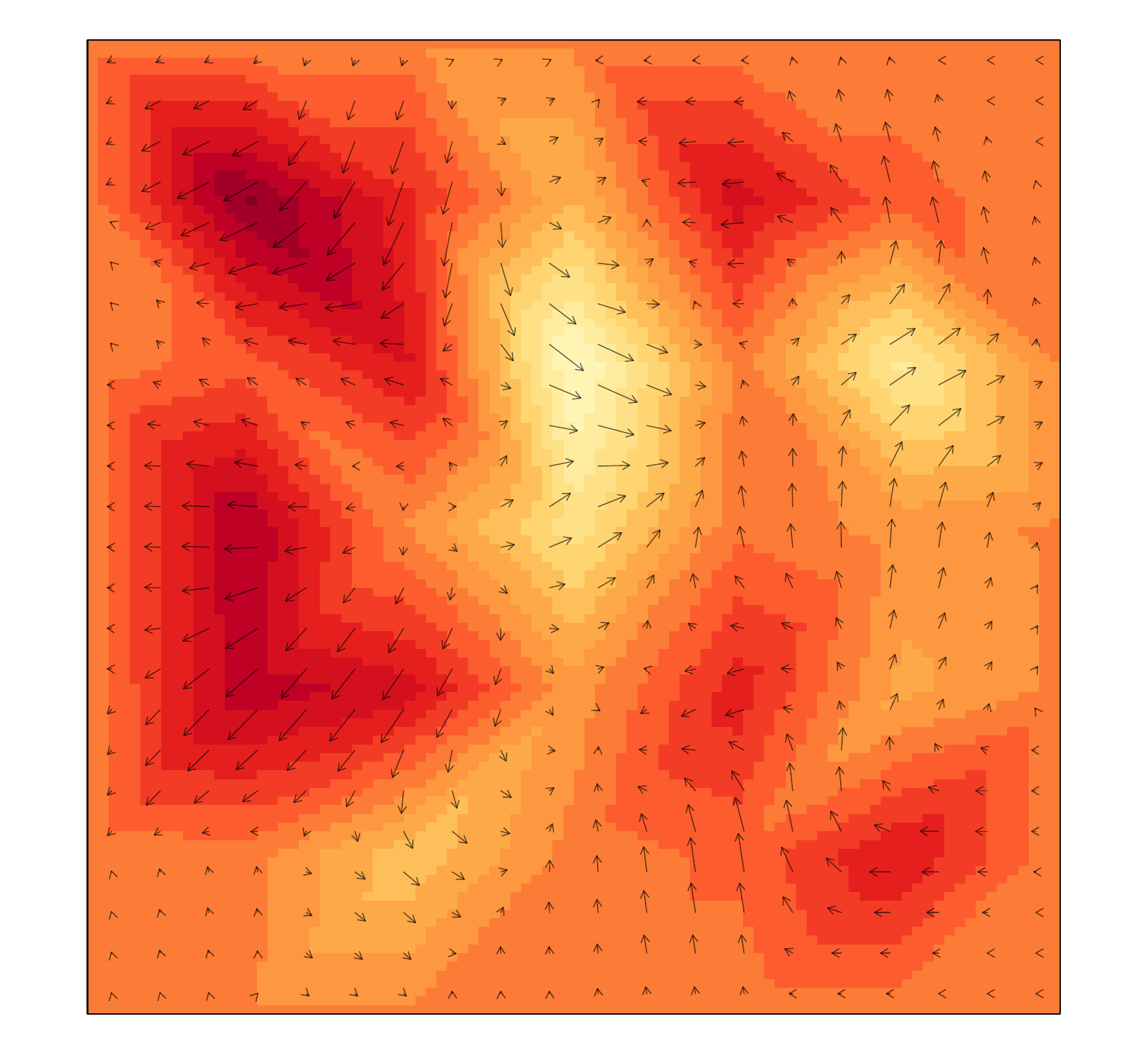}
\caption{An example of disparity vectors at a $5\times 5$ grid of anchor points and the corresponding warping function.}
\label{warpEx}
\end{figure}

We will consider warping functions of the form
\[
v_i(s,t) = v(s,t,\bw_i)= \begin{pmatrix}
s \\ t
\end{pmatrix} + \mathcal{E}_{\bw_i}(s,t),
\]
where $\mathcal{E}_{\bw_i}\colon\RR^2\to\RR^2$ is coordinate-wise bilinear spline interpolation of $\bw_i\in\RR^{m_w^1\times m_w^2\times 2}$ on a lattice spanned by $\boldsymbol{s}_w\in\RR^{m^1_w},\boldsymbol{t}_w\in\RR^{m_w^2}$.
 In other words, $\bw_i$ models discrete spatial displacements at the lattice anchor points. Figure~\ref{warpEx} shows an example of disparity vectors on a grid of anchor points and the corresponding warping function.

 The displacements are modeled as random effects, $\bw_i\sim\mathcal{N}(0,\sigma^2C)$ where $C$ is a $2m_w^1m_w^2\times 2m_w^1m_w^2$ covariance matrix, and, as a result, the warping functions can be considered nonlinear functional random effects. As $\bw_i$ is assumed to be normally distributed with mean zero, small displacements are favorited and hence the warp effect will be less prone to fold. The model is a spatial extension of the phase and amplitude varying population pattern (pavpop) model for curves \cite{RaketSommerMarkussen, pavpop}.

\subsection{Estimation}
\label{sec:est}
First, we will consider estimation of the template $\theta$ from the functional observations, and we will estimate the contributions of the different sources of variation. In the proposed model, this is equivalent to estimating the covariance structure $C$ for the warping parameters, the covariance structure $S$ for the spatially correlated intensity variation, and the noise variance $\sigma^2$. The estimate of the template is found by considering model \eqref{eq:model} in the back-warped template space
\begin{align}
y_i(v_i^{-1}(s_j,t_k))=\theta(s_j,t_k) + x_i(v_i^{-1}(s_j,t_k)) + \tilde{\varepsilon}_{ijk}.
\end{align}
Because every back-warped image represents $\theta$ on the observation lattice, a computationally attractive parametrization is to model $\theta$ using one parameter per observation point, and evaluate non-observation points using bilinear interpolation. This parametrization is attractive, because Henderson's mixed-model equations \cite{henderson1950estimation, robinson1991blup} suggests that the conditional estimate for $\theta(s_j,t_k)$ given $\bw_1,\dots, \bw_n$ is the pointwise average
\begin{align}
\hat{\theta}(s_j,t_k)=\frac{1}{n}\sum_{i=1}^n y_i(v_i^{-1}(s_j,t_k)),
\label{eq:hattheta}
\end{align}
if we ignore the slight change in covariance resulting from the back-warping of the random intensity effects.  As this estimator depends on the warping parameters, the estimation of $\theta$ and the variance parameters has to be performed simultaneously with the prediction of the warping parameters. We note that, as in any linear model, the estimate of the template is generally quite robust against slight misspecifications of the covariance structure. And the idea of estimating the template conditional on the posterior warp is similar to the idea of using a hard EM algorithm for computing the maximum likelihood estimator for $\theta$~\cite{neal1998view}.

We use maximum-likelihood estimation to estimate variance parameters, that is, we need to minimize the negative log-likelihood function of model~\eqref{eq:model}. Note that \eqref{eq:model} contains nonlinear random effects due to the term $\theta(v_i(s,t,\bw_i))$ where $\theta\circ v_i$ is a nonlinear transformation of $\bw_i$.  We handle the nonlinearity and approximate the likelihood by linearizing the model~\eqref{eq:model} around the current predictions $\bw^0_i$ of the warping parameters $\bw_i$:
\begin{align}
\nonumber y_i(s_j,t_k)&\approx\theta(v(s_j,t_k,\bw^0_i)) \\
\nonumber & + (\nabla\theta(v(s_j,t_k,\bw_i^0)))^\top J_{\bw_i}v(s_j,t_k,\bw_i)\Bigr|_{\bw_i = \bw_i^0}(\bw_i - \bw_i^0) \\
\nonumber & + x_i(s_j,t_k) + \varepsilon_{ijk} \\
&= \theta(v(s_j,t_k,\bw_i^0)) + Z_{ijk}(\bw_i - \bw_i^0) + x_i(s_j,t_k) + \varepsilon_{ijk},
\end{align}
where $J_{\bw_i}v(s_j,t_k,\bw_i)$ denotes the Jacobian matrix of $v$ with respect to $\bw_i$ and
\begin{align}
Z_{ijk} =  (\nabla\theta(v(s_j,t_k,\bw_i^0)))^\top J_{\bw_i}v(s_j,t_k,\bw_i)\Bigr|_{\bw_i = \bw_i^0}.
\end{align}
Letting $Z_i=(Z_{ijk})_{jk}\in\RR^{m\times 2m_w^1m_w^2}$, the linearized model can be rewritten
\begin{align}
\boldsymbol{y}_i\approx \boldsymbol{\theta}^{\bw_i^0} + Z_i(\bw_i - \bw_i^0) + \boldsymbol{x}_i + \boldsymbol{\varepsilon_i}.\label{eq:model_lin_vec}
\end{align}
We notice that in this manner, $\boldsymbol{y}_i$ can be approximated as a linear combination of normally distributed variables, hence the negative log-likelihood for the linearized model is given by
\begin{align}
\nonumber\ell_{\boldsymbol{y}}(\theta,C,\sigma^2) = & \frac{nm_1m_2}{2}\log\sigma^2 + \frac{1}{2}\sum_{i=1}^n\log \det V_i \\
&+ \frac{1}{2\sigma^2}\sum_{i=1}^n (\boldsymbol{y}_i - \boldsymbol{\theta}^{\bw_i^0} + Z_i\bw_i^0)^\top V_i^{-1}(\boldsymbol{y}_i - \boldsymbol{\theta}^{\bw_i^0}+ Z_i\bw_i^0),
\label{eq:like}
\end{align}
where $V_i = Z_iCZ_i^\top + S + \II_{m}$. The idea of linearizing nonlinear mixed-effects models in the nonlinear random effects is a solution that has been shown to be effective and which is implemented in standard software packages \cite{LindstromBates, pinheiro2006mixed, pinheiro2007linear}. The proposed model is, however, both more general and computationally demanding than what can be handled by conventional software packages.  Furthermore, we note that the linearization in a random effect as done in model~\eqref{eq:model_lin_vec} is fundamentally different than the conventional linearization of a nonlinear dissimilarity measure such as in the variational problem~\eqref{eq:varprob}. As we see from the linearized model~\eqref{eq:model_lin_vec}, the density of $\theta(v(s_j,t_k,\bw_i)$ is approximated by the density of a linear combination, $\theta(v(s_j,t_k,\bw_i^0)) + Z_{ijk}(\bw_i - \bw_i^0)$, of multivariate Gaussian variables. The likelihood function for the first-order Taylor expansion in $\bw_i$ of the model~\eqref{eq:model} is thus a Laplace approximation of the true likelihood, and the quality of this approximation is approximately second order \cite{wolfinger1993laplace}.

\subsubsection{Computing the likelihood function}\label{sec:computations}
As mentioned above the proposed model is computationally demanding. Even the approximated likelihood function given in equation~\eqref{eq:like} is not directly computable because of the large data sizes. In particular, the computations related to determinants and inverses of the covariance matrix $V_i$ are infeasible unless we impose certain structures on these. In the following, we will assume that the covariance matrix for the spatially correlated intensity variation $S$ has full rank and sparse inverse. We stress that this assumption is merely made for computational convenience and that the proposed methodology is also valid for non-sparse precision matrices. The zeros in the precision matrix $S^{-1}$ are equivalent to assuming conditional independences between the intensity variation in corresponding pixels given all other pixels \cite{lauritzen1996graphical}. A variety of classical models have this structure, in particular (higher-order) Gaussian Markov random fields models have sparse precision matrices because of their Markov property. 

To efficiently do computations with the covariances $V_i=Z_iCZ_i^\top + S + \II_{m}$, we exploit the structure of the matrix. The first term $Z_iCZ_i^\top $ is an update to the intensity covariance $S + \II_{m}$ with a maximal rank of $2m_w^1m_w^2$. Furthermore, the first term of the intensity covariance $S$ has a sparse inverse and the second term $\II_m$ is of course sparse with a sparse inverse. Using the Woodbury matrix identity, we obtain
\begin{align*}
V_i^{-1} &= (Z_iCZ_i^\top + S + \II_m)^{-1} \\
	 &= (S+\II_m)^{-1}-(S+\II_m)^{-1}Z_i(C^{-1}+ Z_i^\top (S+\II_m)^{-1} Z_i)^{-1}Z_i^\top (S+\II_m)^{-1}
\end{align*}
which can be computed if we can efficiently compute the inverse of the potentially huge $m\times m$ intensity covariance matrix $(S+\II_m)^{-1}$. We can rewrite the inverse intensity covariance as
\[
(S+\II_m)^{-1} = \II_m - (\II_m+S^{-1})^{-1}.
\]
Thus we can write $V_i^{-1}$ in a way that only involves operations on sparse matrices. To compute the inner product  $\by^\top V_i^{-1}\by$, we first form the matrix $\II_m+S^{-1}$ and compute its Cholesky decomposition using the Ng-Peyton method \cite{ng1993block} implemented in the \texttt{spam} R-package \cite{furrer2010spam}. By solving a low-rank linear system using the Cholesky decomposition, we can thus compute $L =(C^{-1}+ Z_i^\top (S+\II_m)^{-1} Z_i)^{-1}$. The inner product is then efficiently computed as
\[
\by^\top V_i^{-1}\by = \by^\top \bx - (Z_i \bx)^\top LZ_i \bx
\]
where 
\[
\bx = (S+\II_m)^{-1}\by.
\]
To compute the log determinant in the likelihood, one can use the matrix determinant lemma similarly to what was done above to split the computations into low-rank computations and computing the determinant of $S + \II_m$,
\[
\det(V_i)= \det(Z_i C Z_i^\top+S+\II_{m})= \det(C^{-1}+ Z_i^\top(S+\II_{m})^{-1}Z_i )\det(C)\det(S+\II_{m}).
\]
 For the models that we will consider, the latter computation is done by using the operator approximation proposed in \cite{RaketMarkussen} which, for image data with sufficiently high resolution (e.g. $m > 30$), gives a high-quality approximation of the determinant of the intensity covariance that can be computed in constant time. 

By taking the described strategy, we never need to form a dense $m\times m$ matrix, and we can take advantage of the sparse and low-rank structures to reduce the computation time drastically. Furthermore, the fact that we assume equal-size images allows us to only do a single Cholesky factorization per likelihood computation, which is further accelerated by using the updating scheme described in \cite{ng1993block}.

\subsection{Prediction}
\label{sec:pred}

After the maximum-likelihood estimation of the template $\theta$ and the variance parameters, we have an estimate for the
distribution of the warping parameters. We are therefore able to predict the
warping functions that are most likely to have occurred given the observed data. This prediction parallels the conventional estimation of deformation functions in image registration.
Let $p_{w_i|y_i}$ be the density for the
distribution of the warping functions given the data and define $p_{w_i}$,
$p_{y_i|w_i}$ in a similar manner. Then, by applying
$p_{w_i|y_i}\propto p_{y_i|w_i}p_{w_i}$, we see that the warping functions that are
most likely to occur are the minimizers of the posterior
\begin{align}
-\log(p_{w_i|y_i})\propto  \frac{1}{2\sigma^2}(\boldsymbol{y}_i - \boldsymbol{\theta}^{\bw_i})^\top (S + I_{m})^{-1}(\boldsymbol{y}_i - \boldsymbol{\theta}^{\bw_i}) + \frac{1}{2\sigma^2}\bw_i^\top C^{-1}\bw_i.
\label{eq:wlike}
\end{align} 
 Given the updated predictions $\hat\bw_i$ of the warping parameters, we update the estimate of the template and then minimize the likelihood $(\ref{eq:like})$ to obtain updated estimates of the variances. This procedure is then repeated until convergence is obtained. The estimation algorithm is given in Algorithm~\ref{alg}. The run time for the algorithm will be very different depending on the data in question. As an example we ran the model for 10 MRI midsaggital slices (for more details see Section~\ref{MRI}) of size $210\times 210$, with $i_{\max} = 5, j_{\max} = 3$. We ran the algorithm on an Intel Xeon E5-2680 2.5GHz processor. The run time needed for full maximum likelihood estimation in this setup was 1 hour and 15 minutes using a single core. This run time is without parallization, but it is possible to apply parallization to make the algorithm go faster.

The spatially correlated intensity variation can also be predicted. Either as the best linear unbiased prediction $\mathrm{E}[\bx_i\,|\, \by]$ from the linearized model~\eqref{eq:model_lin_vec} (see e.g. equation~5 in \cite{Markussen}). Alternatively, to avoid a linear correction step when predicting $\bw_i$,  one can compute the best linear unbiased prediction given the maximum-a-posteori warp variables 
\begin{align}
\mathrm{E}[x_i(s, t)\,|\, \by_i, \bw_i = \hat\bw_i]= S (S + I_m)^{-1}(\boldsymbol{y}_i - \boldsymbol{\hat{\theta}}^{\hat\bw_i}).\label{eq:xpred}
\end{align} 
The prediction of the spatially correlated intensity variation can, for example, be used for bias field correction of the images.

\begin{algorithm} \DontPrintSemicolon \SetAlgoLined
\KwData{$\by$} 
\KwResult{Estimates of the fixed effect and variance parameters of the model, and the resulting predictions of the warping parameters $\bw$}  
\tcp{Initialize parameters} 
Initialize $\bw^0$\;
Compute $\hat\btheta^{\bw^0}$ following $(\ref{eq:hattheta})$\;
\For{$i=1$ to $i_{\max}$}{ 
	\tcp{Outer loop: parameters} 
    	Estimate variance parameters by minimizing $(\ref{eq:like})$\;
	\For{$j=1$ to $j_{\max}$}{
      		\tcp{Inner loop: fixed effect, warping parameters}
		Predict warping parameters by minimizing $(\ref{eq:wlike})$\;
		Update linearization points $\bw^0$ to current prediction\;
		Recompute $\hat\btheta^{\bw^0}$ from $(\ref{eq:hattheta})$ 
	}
}
\caption{Inference in the model $(\ref{eq:model})$.}\label{alg}
\end{algorithm}

\section{Models for the spatially correlated variations} 
\label{sec:invB}

The main challenge of the presented methods is the computability of the likelihood function, in particular computations related to the $m\times m $ covariance matrix of the spatially correlated intensity variation $S$. The same issues are not associated with the covariance matrix $C$, for the warping parameters, as the dimensions of this matrix are considerably smaller than the dimensions of $S$. In the end of this section, we will give a short description of how the displacement vectors can be modeled, but first we consider the covariance matrix $S$.

As mentioned in the previous section, the path we will pursue to make likelihood computations efficient is to assume that the systematic random effect $\bx_i$ has a covariance matrix $S$ with sparse inverse. In particular, modeling $\bx_i$ as a Gaussian Markov random field will give sparse precision matrices $S^{-1}$. The Markov random field structure gives a versatile class of models that has been demonstrated to be able to approximate the properties of general Gaussian fields surprisingly well~\cite{rue2002fitting}. Estimation of a sparse precision matrix is a fundamental problem and a vast literature exists on the subject. We mention in passing the fundamental works,~\cite{cai_constrained_2011,friedman_sparse_2008}, which could be adapted to the present setup to estimate unstructured sparse precision matrices. We will however not pursue that extension in the present paper.

We here model $\bx_i$ as a tied-down Brownian sheet, which is the generalization of the Brownian bridge (which is Markov) to the unit square $[0,1]^2$. The covariance function, $\mathcal{S}\colon [0,1]^2\times [0,1]^2\to\mathbb{R}$, for the tied-down Brownian sheet is
\[
\mathcal{S}((s,t), (s', t')) = \tau^2 (s\wedge s' - ss')(t\wedge t' - tt'), \qquad \tau > 0.
\]
The covariance is $0$ along the boundary of the unit square and reaches its maximal variance at the center of the image. These properties seem reasonable for many image analysis tasks, where one would expect the subject matter to be centered in the image with little or no variation along the image boundary.

Let $S$ be the covariance matrix for a Brownian sheet observed at the lattice spanned by $(s_1,\dots, s_{m_1})$ and $(t_1,\dots, t_{m_2})$, $s_i = i / (m_1 + 1)$, $t_i = i / (m_2 + 1)$ with row-major ordering. The precision matrix $S^{-1}$ is sparse with the following structure for points corresponding to non-boundary elements:
\[
\frac{1}{\tau^2(m_1 + 1)(m_2+1)}S^{-1}[i,j] = \begin{cases}
\phantom{-}4  & \textrm{if } j=i\\
-2 & \textrm{if } j\in \{i-1, i+1, i + m_2, i - m_2 \}\\
\phantom{-}1  & \textrm{if } j\in \{i-1 - m_2, i+1 - m_2, i -1 + m_2, i + 1 + m_2 \}
\end{cases}.
\]
For boundary elements, the $j$ elements outside the observation boundary vanish. 

As explained in Section~\ref{sec:computations}, the computational difficulties related to the computation of the log determinant in the negative log likelihood function~\eqref{eq:like} comes down to computing the log determinant of the intensity covariance $S + \II_m$. For the tied-down Brownian sheet, the log determinant can be approximated by means of the operator approximation given in \cite[Example~3.4]{RaketMarkussen}. The approximation is given by
\[
\log\det(S + \II_m) = \sum_{\ell=1}^\infty \log\bigg(\frac{ \pi \ell}{\sqrt{\tau^2 (m_1 + 1)(m_2 + 1)}}\sinh\bigg(\frac{\sqrt{\tau^2 (m_1 + 1)(m_2 + 1)}}{\pi\ell}\bigg)\bigg).
\]
To compute the approximation we cut the sum off after 10,000 terms. 

As a final remark, we note that the covariance function $\tau^{-2}\mathcal{S}$ is the Green's function for the differential operator $\partial_s^2\partial_t^2$ on $[0,1]^2$ under homogeneous Dirichlet boundary conditions. Thus the conditional linear prediction of $\bx_i$ given by \eqref{eq:xpred} is equivalent to estimating the systematic part of the residual as a generalized smoothing spline with roughness penalty 
\[
\frac{1}{2\tau^2}\int_0^1 \int_0^1 x_i(s, t) \partial_s^2\partial_t^2 x_i(s, t)\,\ud s\,\ud t =\frac{1}{2\tau^2} \int_0^1 \int_0^1 \|\partial_s\partial_t x_i(s, t)\|^2\,\ud s\,\ud t.
\]

The tied-down Brownian sheet can also be used to model the covariance between the displacement vectors. Here the displacement vectors given by the warping variables $\bw_i$ are modeled as discretely observed tied-down Brownian sheets in each displacement coordinate. As was the case for the intensity covariance, this model is a good match to image data since it allows the largest deformations around the middle of the image. Furthermore, the fact that the model is tied down along the boundary means that we will predict the warping functions to be the identity along the boundary of the domain $[0, 1]^2$, and for the found variance parameters, the predicted warping functions will be homeomorphic maps of $[0,1]^2$ onto $[0,1]^2$ with high probability.

In the applications in the next section, we will use the tied-down Brownian sheet to model the spatially correlated variations.

\section{Applications}
\label{sec:app}

In this section, we will apply the developed methodology on two different real-life datasets. In the first example, we apply the model to a collection of face images that are difficult to compare due to varying expressions and lighting sources. We compare the results of the proposed model to conventional registration methods and demonstrate the effects of the simultaneous modeling of intensity and warp effects. In the second example, we apply the methodology to the problem of estimating a template from affinely aligned 2D MR images of brains. 

\subsection{Face registration}
\begin{figure}[!th]
\includegraphics[width = 0.19\textwidth, trim = 30 0 30 0, clip]{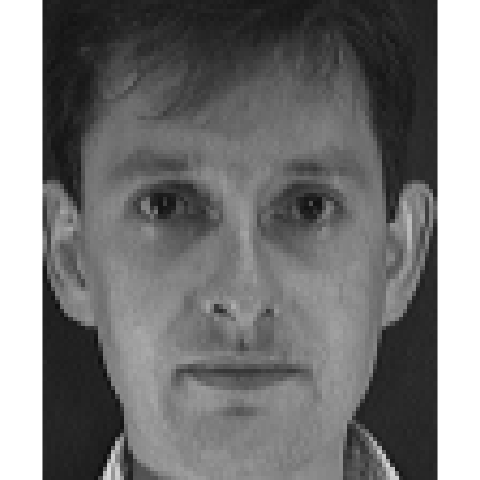}
\includegraphics[width = 0.19\textwidth, trim = 30 0 30 0, clip]{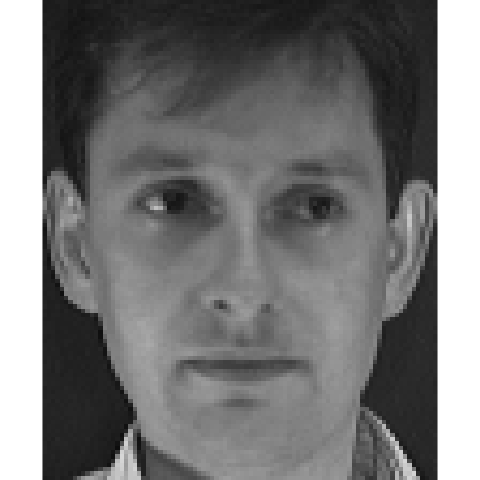}
\includegraphics[width = 0.19\textwidth, trim = 30 0 30 0, clip]{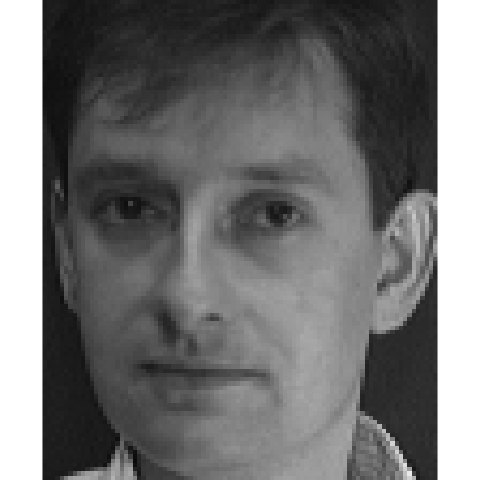}
\includegraphics[width = 0.19\textwidth, trim = 30 0 30 0, clip]{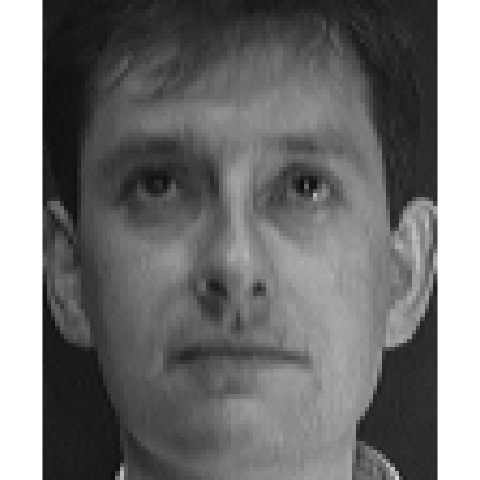}
\includegraphics[width = 0.19\textwidth, trim = 30 0 30 0, clip]{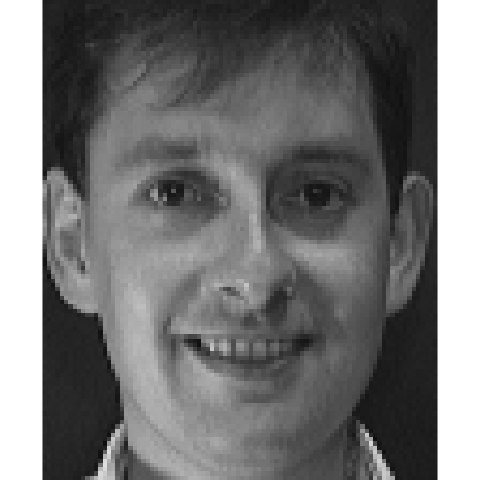}
\\[0.5em]
\includegraphics[width = 0.19\textwidth, trim = 30 0 30 0, clip]{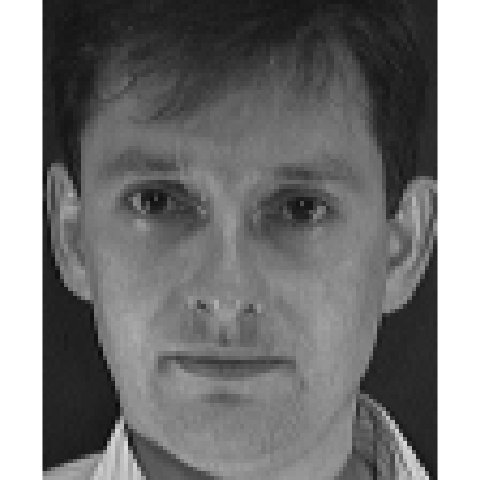}
\includegraphics[width = 0.19\textwidth, trim = 30 0 30 0, clip]{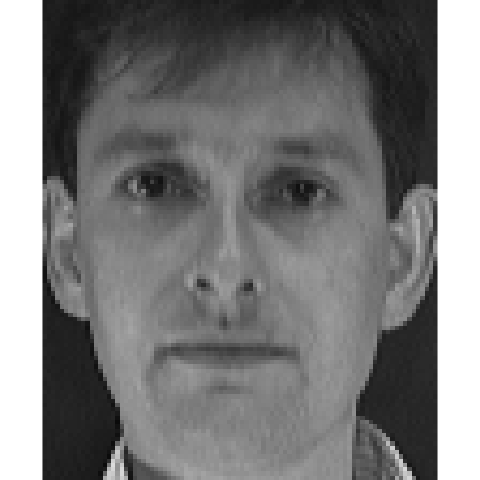}
\includegraphics[width = 0.19\textwidth, trim = 30 0 30 0, clip]{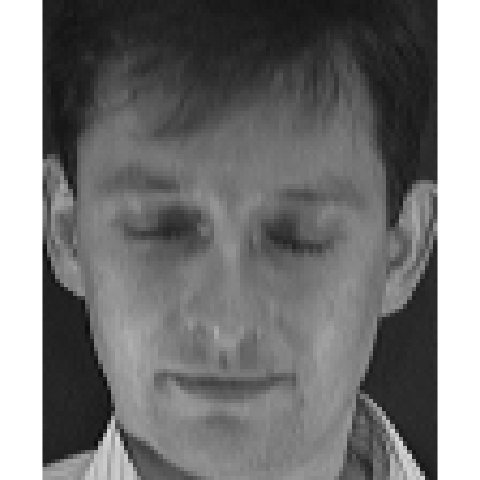}
\includegraphics[width = 0.19\textwidth, trim = 30 0 30 0, clip]{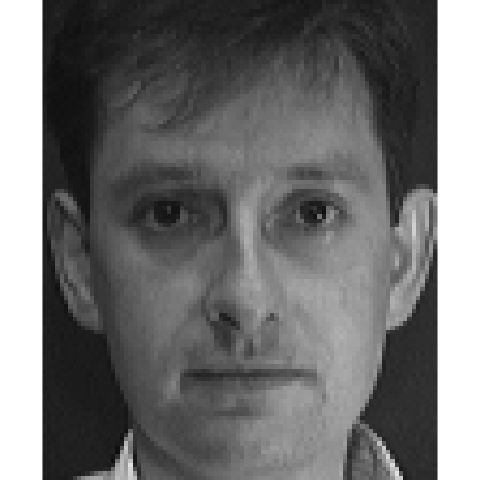}
\includegraphics[width = 0.19\textwidth, trim = 30 0 30 0, clip]{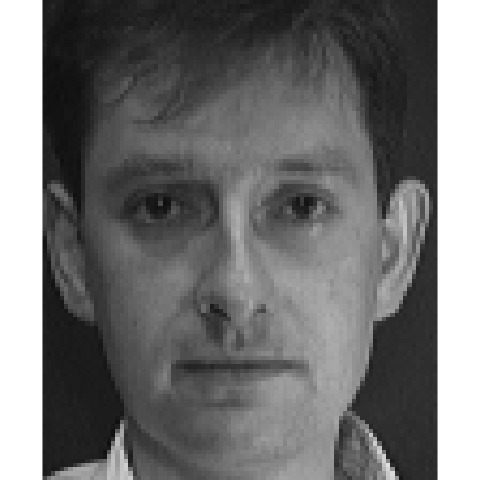}
\caption{Ten images of the same face with varying expressions and illumination. The images are from the AT\&T Laboratories Cambridge Face Database~\cite{att}.}\label{fig:faces1}
\end{figure}
Consider the ten  $92\times 112$ face images from the AT\&T Laboratories Cambridge Face Database~\cite{att} in Figure~\ref{fig:faces1}. The images are all of the same person, but vary in head position, expression and lighting. The dataset contains two challenges from a registration perspective, namely the differences in expression that cause dis-occlusions or occlusions (e.g. showing teeth, closing eyes) resulting in large local deviations; and the difference in placement of the lighting source that causes strong systematic deviations throughout the face. 

To estimate a template face from these images, the characteristic features of the face should be aligned, and the systematic local and global deviations should be accounted for. In the proposed model~\eqref{eq:model}, these deviations are explicitly modeled through the random effect $x_i$.

Using the maximum-likelihood estimation procedure, we fitted the model to the data using displacement vectors $\bw_i$ on an equidistant $4\times 4$ interior grid in $[0,1]^2$. We used 5 outer and 3 inner iterations in Algorithm~\ref{alg}.  The image value range was scaled to $[0,1]$. The estimated variance scale for the random effect $x_i$ was $\hat\sigma^2\hat\tau^2 = 0.658$; for the warp variables, the variance scale was estimated to $\hat\sigma^2\hat\gamma^2 = 0.0680$; and for the residual variance, the estimated scale was $\hat\sigma^2=0.00134$.

To illustrate the effect of the simultaneous modeling of random intensity and warp effects, we estimated a face template using three more conventional variants of the proposed framework: a pointwise estimation that corresponds to model~\eqref{eq:model} with no warping effect; a \emph{Procrustes} model that corresponds to model~\eqref{eq:model} with no intensity component and where the warp variables $\bw_i$ were modeled as unknown parameters and estimated using maximum-likelihood estimation; and a \emph{warp-regularized Procrustes} method where the warp variables $\bw_i$ were penalized using a term $\lambda \bw_i^\top C^{-1} \bw_i$ where $C^{-1}$ is the precision matrix for the 2D tied-down Brownian sheet with smoothing parameter $\lambda = 3.125$ (chosen to give good visual results).

\begin{figure}[!pt]
\centering
\begin{tabular}{p{0.22\textwidth}p{0.22\textwidth}p{0.22\textwidth}p{0.22\textwidth}}
\centering\scriptsize No alignment & \centering\scriptsize Procrustes free warp & \centering\scriptsize{Procrustes regularized warp} & \centering\scriptsize proposed   \cr
\centering\includegraphics[width = 0.2\textwidth, trim = 30 0 30 0, clip]{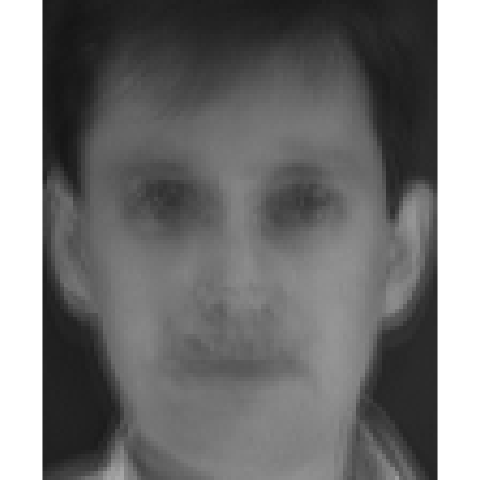} &
\centering\includegraphics[width = 0.2\textwidth, trim = 30 0 30 0, clip]{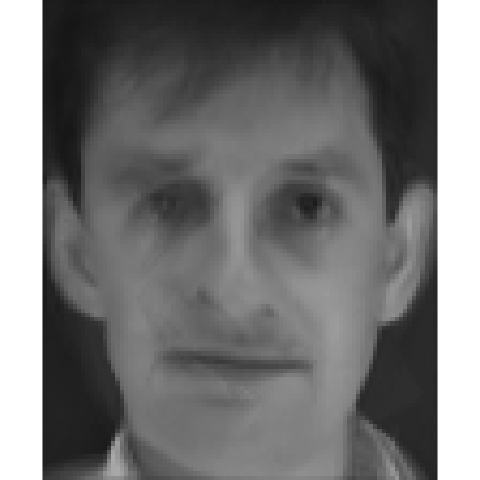} &
\centering\includegraphics[width = 0.2\textwidth, trim = 30 0 30 0, clip]{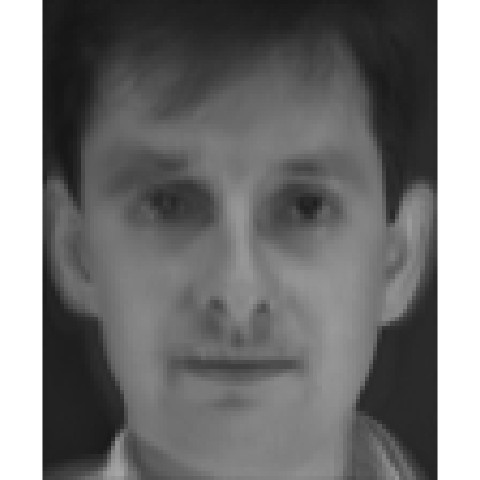} &
\centering\includegraphics[width = 0.2\textwidth, trim = 30 0 30 0, clip]{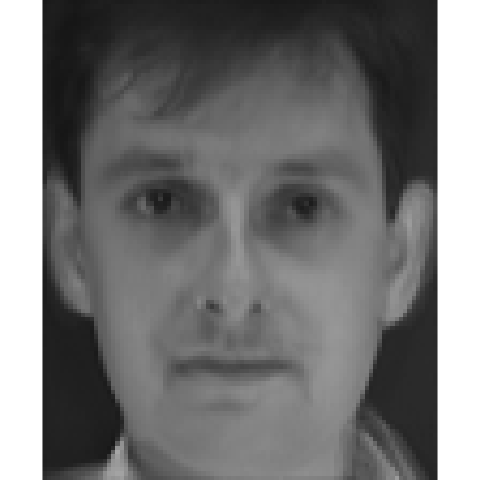}
\end{tabular}
\caption{Estimates for the fixed effect $\theta$ using different models. The models used to calculate the estimates are from left to right: model assuming no warping effect and Gaussian white noise for the intensity model, the same model but with a free warping function based on 16 displacement vectors, the same model but with a penalized estimation of warping functions (2D tied-down Brownian sheet with scale fixed $\tau=0.4$), the full model~\eqref{eq:model}.}\label{fig:faces2}
\end{figure}

The estimated templates for the proposed model and the alternative models described above can be found in Figure~\ref{fig:faces2}. Going from left to right, it is clear that the sharpness and representativeness of the estimates increase. 

\begin{figure}[!pt]
\centering
\begin{tabular}{p{0.15\textwidth}|p{0.15\textwidth}|p{0.15\textwidth}|p{0.15\textwidth}|p{0.15\textwidth}}
\centering\scriptsize Procrustes & \centering\scriptsize{Regularized Procrustes} & \centering\scriptsize proposed warped template prediction & \centering\scriptsize proposed full prediction & \centering\scriptsize observation \cr
\centering\includegraphics[width = 0.15\textwidth, trim = 30 0 30 0, clip]{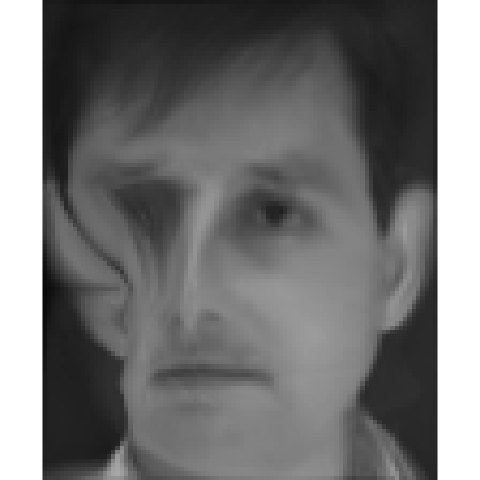} &
\centering\includegraphics[width = 0.15\textwidth, trim = 30 0 30 0, clip]{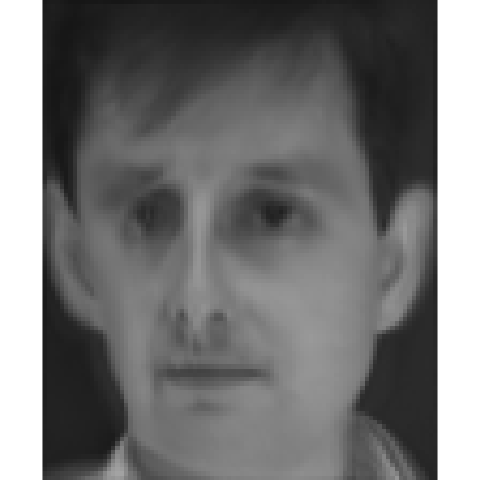} &
\centering\includegraphics[width = 0.15\textwidth, trim = 30 0 30 0, clip]{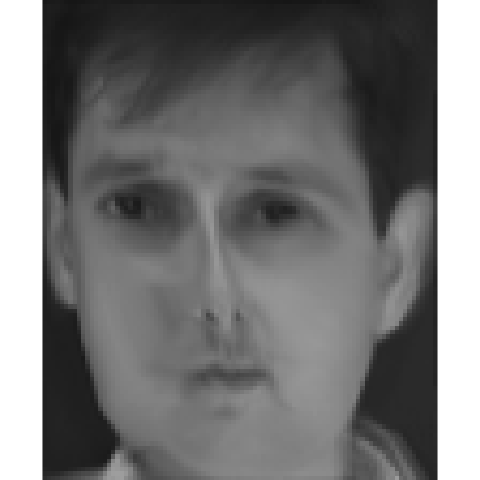} &
\centering\includegraphics[width = 0.15\textwidth, trim = 30 0 30 0, clip]{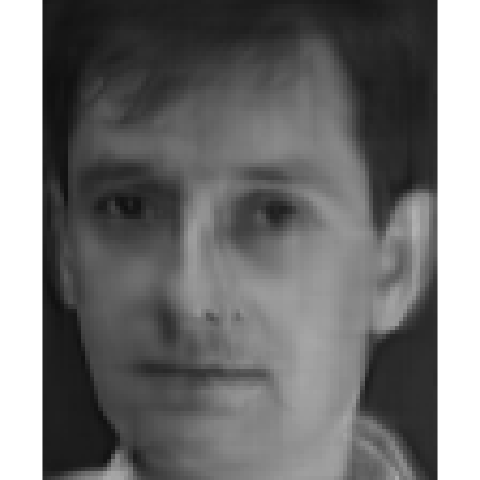} &
\centering\includegraphics[width = 0.15\textwidth, trim = 30 0 30 0, clip]{img/y3.png} \cr
\centering\includegraphics[width = 0.15\textwidth, trim = 30 0 30 0, clip]{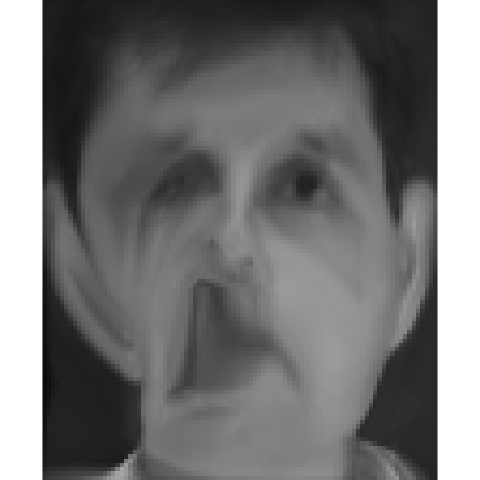} &
\centering\includegraphics[width = 0.15\textwidth, trim = 30 0 30 0, clip]{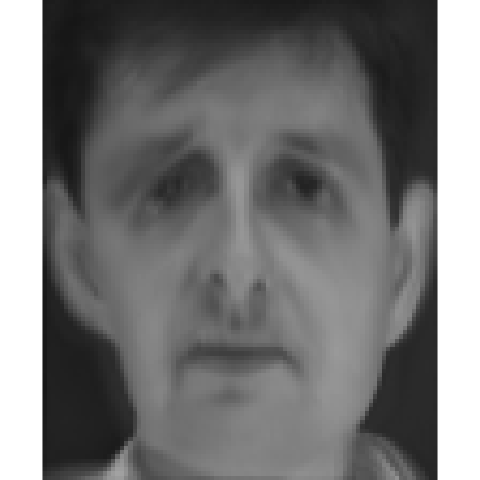} &
\centering\includegraphics[width = 0.15\textwidth, trim = 30 0 30 0, clip]{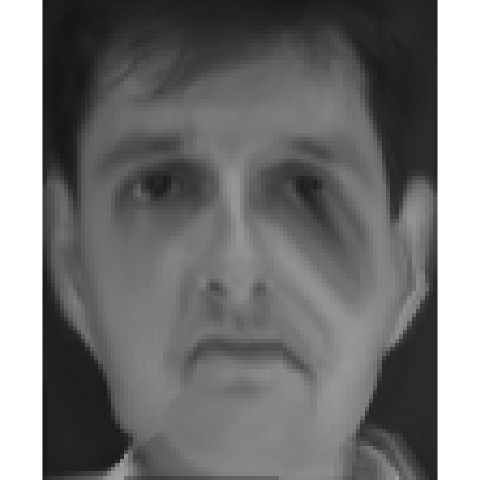} &
\centering\includegraphics[width = 0.15\textwidth, trim = 30 0 30 0, clip]{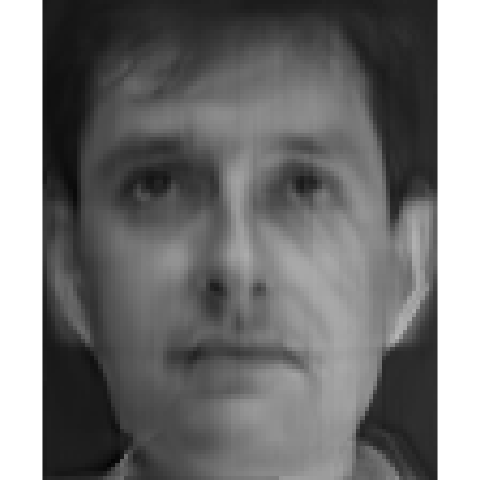} &
\centering\includegraphics[width = 0.15\textwidth, trim = 30 0 30 0, clip]{img/y4.png} \cr
\centering\includegraphics[width = 0.15\textwidth, trim = 30 0 30 0, clip]{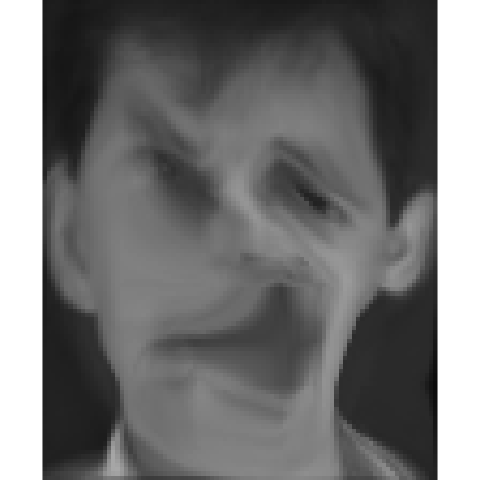} &
\centering\includegraphics[width = 0.15\textwidth, trim = 30 0 30 0, clip]{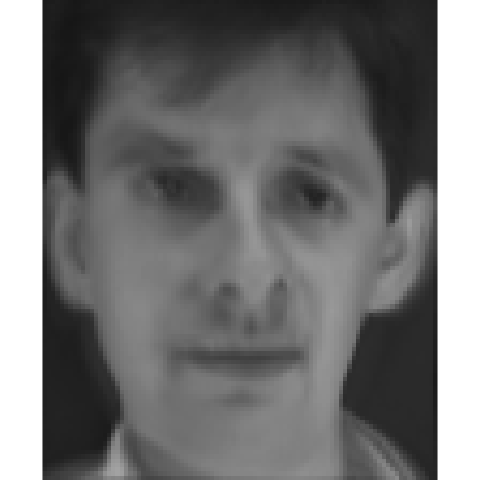} &
\centering\includegraphics[width = 0.15\textwidth, trim = 30 0 30 0, clip]{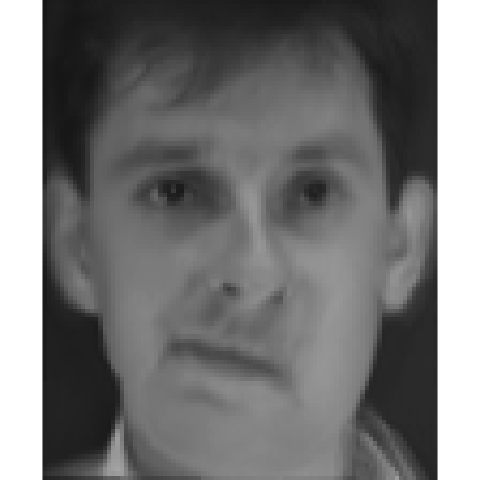} &
\centering\includegraphics[width = 0.15\textwidth, trim = 30 0 30 0, clip]{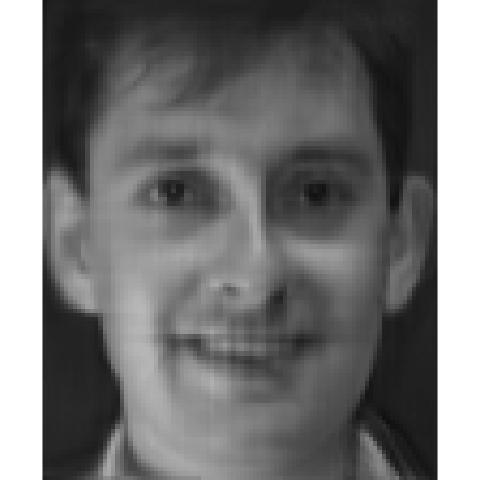} &
\centering\includegraphics[width = 0.15\textwidth, trim = 30 0 30 0, clip]{img/y5.png} \cr
\centering\includegraphics[width = 0.15\textwidth, trim = 30 0 30 0, clip]{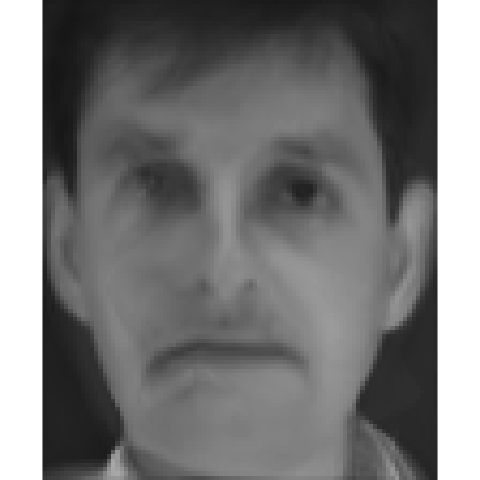} &
\centering\includegraphics[width = 0.15\textwidth, trim = 30 0 30 0, clip]{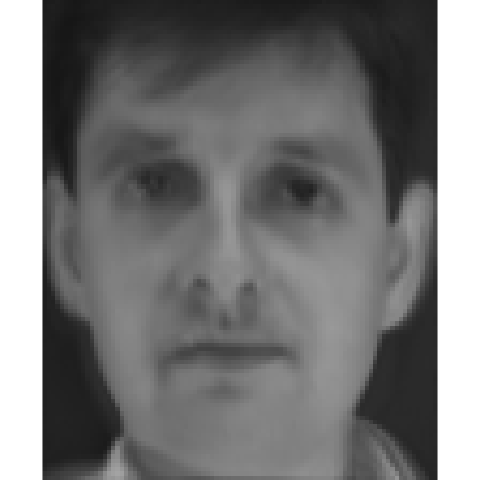} &
\centering\includegraphics[width = 0.15\textwidth, trim = 30 0 30 0, clip]{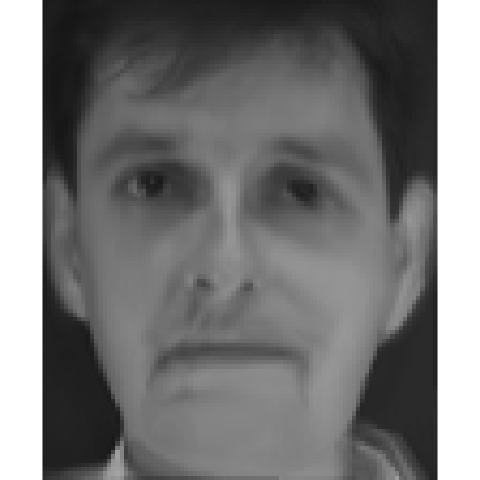} &
\centering\includegraphics[width = 0.15\textwidth, trim = 30 0 30 0, clip]{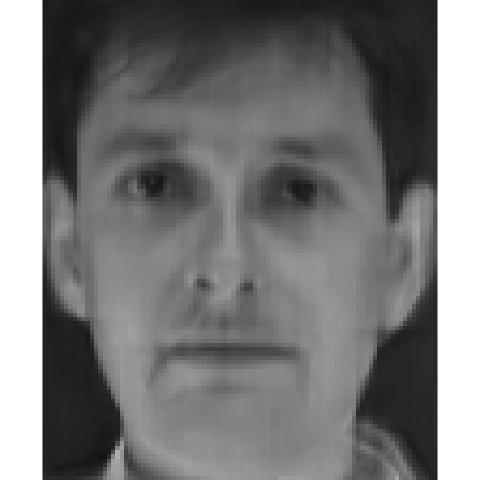} &
\centering\includegraphics[width = 0.15\textwidth, trim = 30 0 30 0, clip]{img/y7.png} \cr
\centering\includegraphics[width = 0.15\textwidth, trim = 30 0 30 0, clip]{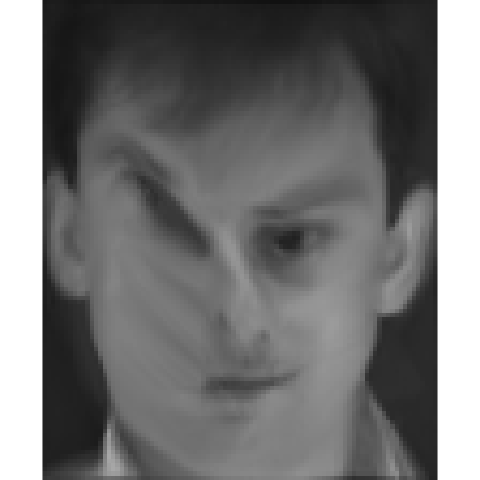} &
\centering\includegraphics[width = 0.15\textwidth, trim = 30 0 30 0, clip]{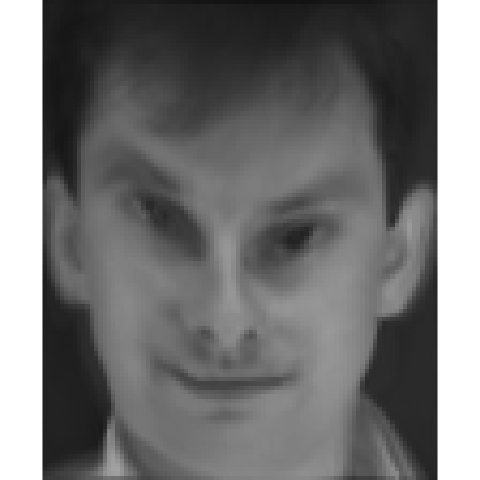} &
\centering\includegraphics[width = 0.15\textwidth, trim = 30 0 30 0, clip]{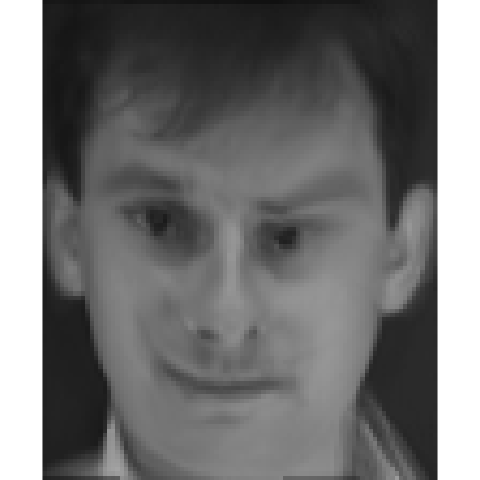} &
\centering\includegraphics[width = 0.15\textwidth, trim = 30 0 30 0, clip]{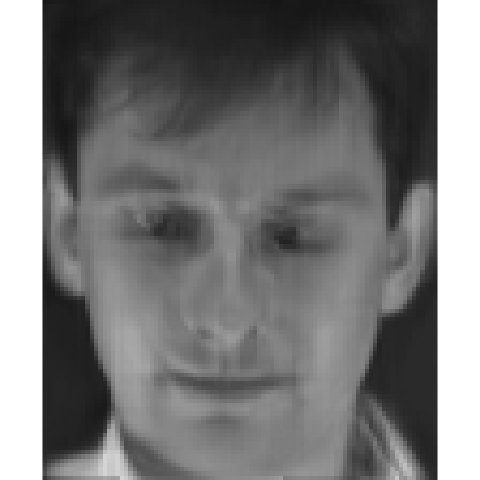} &
\centering\includegraphics[width = 0.15\textwidth, trim = 30 0 30 0, clip]{img/y8.png}
\end{tabular}
\caption{Model predictions of five face images (rightmost column). The two first columns display the maximum-likelihood predictions from the Procrustes and regularized Procrustes models. The third column displays the warped template $\hat\theta(v(s, t, \hat\bw_i))$ where $\hat\bw_i$ is the most likely warp given data $\by$. The fourth column displays the full conditional prediction given the posterior warp variables $\hat\theta(v(s, t, \hat\bw_i)) + \mathrm{E}[x_i(s, t)\,|\, \by_i, \bw_i = \hat\bw_i]$.}\label{fig:faces3}
\end{figure}

To validate the models, we can consider how well they predict the observed faces under the maximum-likelihood estimates and posterior warp predictions. These predictions are displayed in Figure~\ref{fig:faces3}. The rightmost column displays the five most deviating observed faces. From the left, the first three columns show the corresponding predictions from the Procrustes model, the warp-regularized Procrustes model and, for comparison, the predicted warped templates from the proposed model. It is clear that both the sharpness and the representativeness increase from left to right. The predictions in the third column show the warped template of model~\eqref{eq:model} which does not include the predicted intensity effect $x_i$. The fourth column displays the full prediction from the proposed model given as the best linear unbiased prediction conditional on the maximum-a-posteori warp variables $\hat\theta(v(s, t, \hat\bw_i)) + \mathrm{E}[x_i(s, t)\,|\, \by_i, \bw_i = \hat\bw_i]$. The full predictions are very faithful to the observations, with only minor visible deviations around the eyes in the second and fifth row.  This suggests that the chosen model for the spatially correlated intensity variation, the tied-down Brownian sheet, is sufficiently versatile to model the systematic part of the residuals.

\subsection{MRI slices}
\label{MRI}
\begin{figure}[!th]
\centering
\includegraphics[width = 0.30\textwidth, trim = 60 70 30 60, clip]{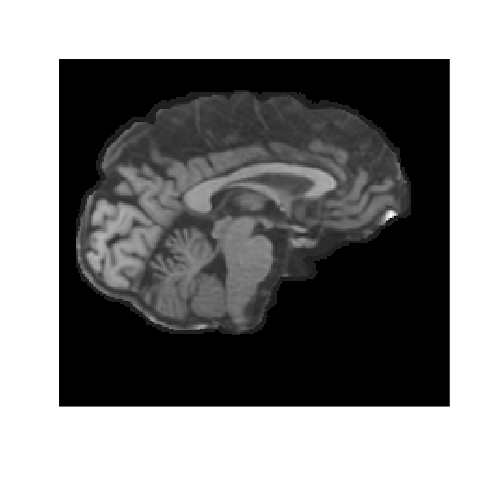}
\includegraphics[width = 0.30\textwidth, trim = 60 70 30 60, clip]{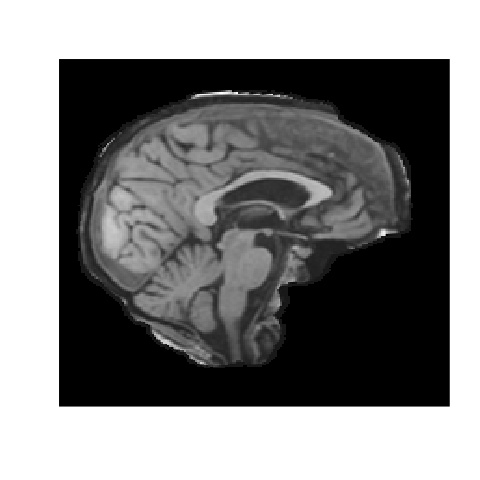}
\includegraphics[width = 0.30\textwidth, trim = 60 70 30 60, clip]{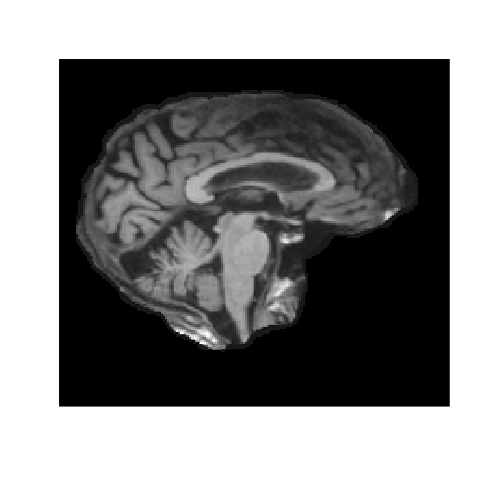}
\\[0.2em]
\includegraphics[width = 0.30\textwidth, trim = 60 70 30 60, clip]{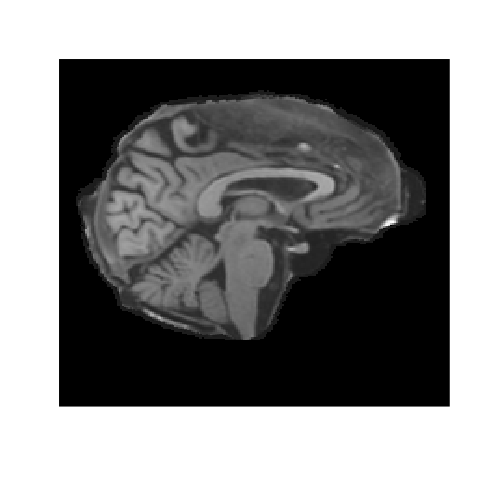}
\includegraphics[width = 0.30\textwidth, trim = 60 70 30 60, clip]{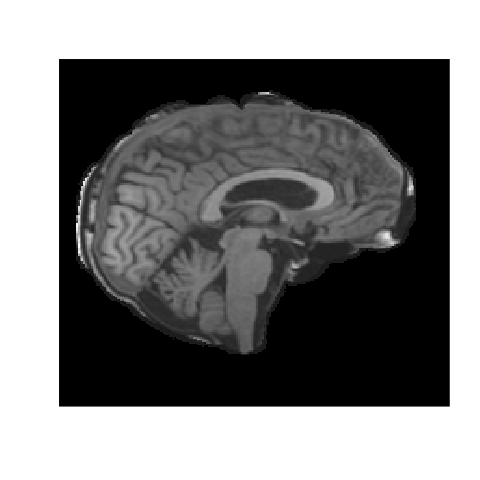}
\includegraphics[width = 0.30\textwidth, trim = 60 70 30 60, clip]{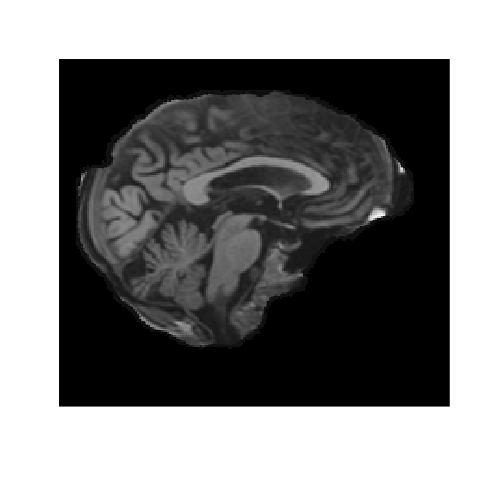}
\caption{A sample of six MRI slices from the data set of 50 mid-sagittal MRI slices.}\label{fig:EksBrains}
\end{figure}

The data considered in this section are based on 3D MR images from the ADNI database~\cite{pai_kernel_2015}. We have based the example on $50$ images with 18 normal controls (NC), 13 with Alzheimer's disease (AD) and 19 who are mild cognitively impaired (MCI). The 3D images were initially affinely aligned with 12 degrees of freedom and normalized mutual information (NMI) as a similarity measure. After the registration, the mid-sagittal slices were chosen as observations. Moreover the images were intensity normalized to $[0,1]$ and afterwards the mid-sagittal plane was chosen as the final observations. The $50$ mid-sagittal planes are given as $210\times 210$ observations on an equidistant grid on $[0,1]^2$. Six samples are displayed in Figure~\ref{fig:EksBrains} where differences in both contrast, placement and shape of the brains are apparent. 

\begin{figure}[!pt]
\centering
\begin{tabular}{p{0.31\textwidth}p{0.31\textwidth}p{0.31\textwidth}}
\centering\scriptsize Rigid registration with scaling & \centering\scriptsize Procrustes free warp & \centering\scriptsize{proposed} \cr
\centering\includegraphics[width = 0.31\textwidth, trim = 60 70 30 60, clip]{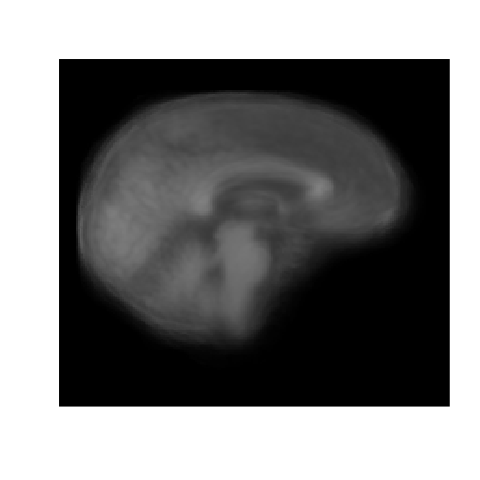} &
\centering\includegraphics[width = 0.31\textwidth, trim = 60 70 30 60, clip]{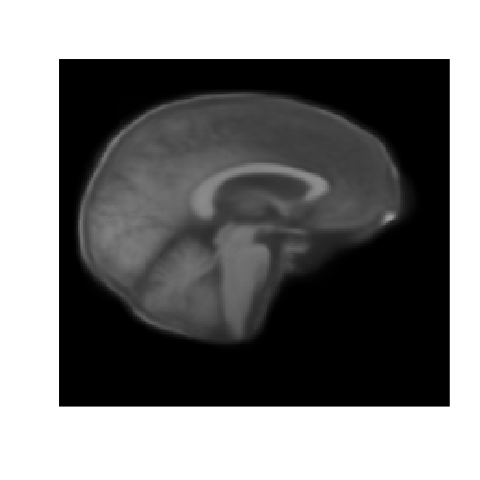} &
\centering\includegraphics[width = 0.31\textwidth, trim = 60 70 30 60, clip]{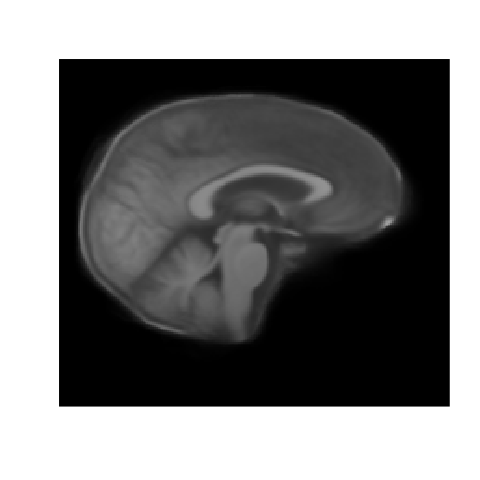} 
\end{tabular}
\caption{Estimates for the fixed effect $\theta$ in three different models. From left to right: pointwise mean after rigid registration and scaling; non-regularized Procrustes; and the proposed model~\eqref{eq:model}.}\label{fig:temp}
\end{figure}

For the given data, we used 25 displacement vectors $\boldsymbol{w}_i$ on an equidistant $5\times 5$ interior grid in $[0,1]^2$. The number of inner iterations in the algorithm was set to $3$, while the number of outer iterations was set to $5$ as the variance parameters and likelihood value already stabilized after a couple of iterations. The estimated variance scales are given by $\hat{\sigma}^2\hat{\tau}^2 = 2.23$ for the spatially correlated intensity variation, $\hat{\sigma}^2\hat{\gamma}^2 = 0.202$ for the warp variation and $\hat{\sigma}^2=7.79\cdot 10^{-4}$ for the residual variance. The estimated template can be found in the rightmost column in Figure~\ref{fig:temp}.

For comparison, we have estimated a template without any additional warping (i.e. only using the rigidly aligned slices), and a template estimated using a Procrustes model with fixed warping effects and no systematic intensity variation, but otherwise comparable to the proposed model. These templates can be found in the leftmost and middle columns of Figure~\ref{fig:temp}. Comparing the three, we see a clear increase in details and sharpness from left to right. The reason for the superiority of the proposed method is both that the regularization of warps is based on maximum-likelihood estimation of variance parameters, but also that the prediction of warps takes the systematic deviations into account. Indeed, we can rewrite the data term in the posterior~\eqref{eq:wlike} as
\[
(\boldsymbol{y}_i - \boldsymbol{\theta}^{\bw_i} - \mathrm{E}[\bx_i\,|\, \by_i, \bw_i])^\top(\by_i - \boldsymbol{\theta}^{\bw_i} - \mathrm{E}[\bx_i\,|\, \by_i, \bw_i])+\mathrm{E}[\bx_i\,|\, \by_i, \bw_i]^\top S^{-1} \mathrm{E}[\bx_i\,|\, \by_i, \bw_i].
\]
Thus, in the prediction of warps, there is a trade-off between the regularity of the displacement vectors (the term $\bw_i^\top C^{-1}\bw_i$ in eq. \ref{eq:wlike}) and the regularity of the predicted spatially correlated intensity variation given the displacement vectors (the term $\mathrm{E}[\bx_i\,|\, \by_i, \bw_i]^\top S^{-1} \mathrm{E}[\bx_i\,|\, \by_i, \bw_i]$).

\begin{figure}[!th]
\centering
\includegraphics[width = 0.31\textwidth, trim = 40 15 40 15, clip]{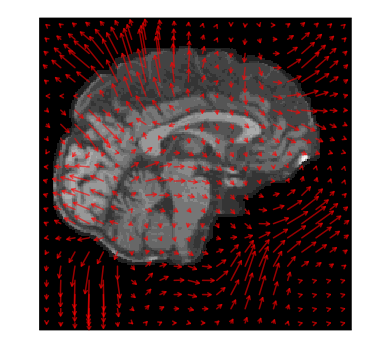} 
\includegraphics[width = 0.31\textwidth, trim = 40 15 40 15, clip]{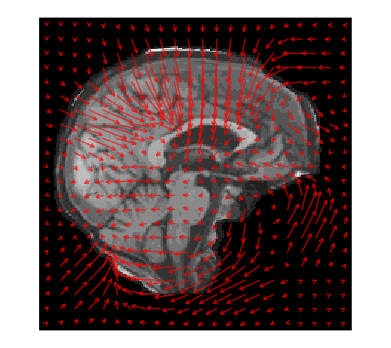} 
\includegraphics[width = 0.31\textwidth, trim = 40 15 40 15, clip]{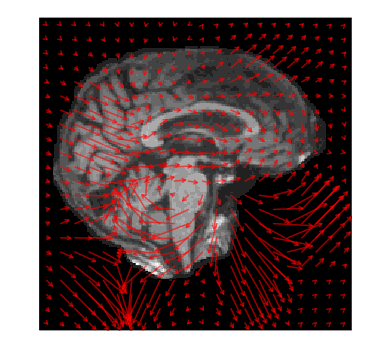} 
\\[0.2em]
\includegraphics[width = 0.31\textwidth, trim = 40 15 40 15, clip]{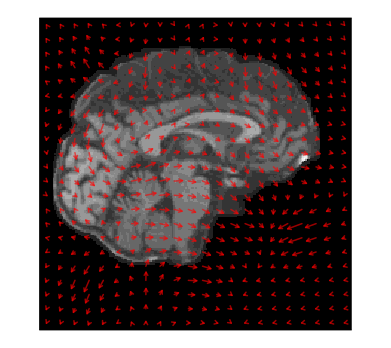} 
\includegraphics[width = 0.31\textwidth, trim = 40 15 40 15, clip]{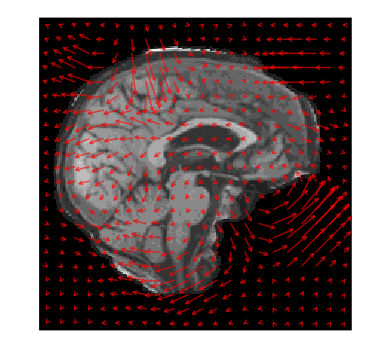} 
\includegraphics[width = 0.31\textwidth, trim = 40 15 40 15, clip]{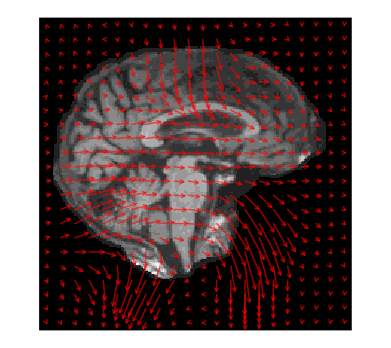} 
\caption{Three MRI slices and their estimated/predicted warping functions for the Procrustes model and the proposed model. The top row shows the Procrustes displacement fields, while the displacement fields for the proposed model are given in the bottom row. The arrows corresponds to the deformation of the observation to the template.}\label{fig:Dis}
\end{figure}

The difference in regularization of the warps is shown in Figure~\ref{fig:Dis}, where the estimated warps using the Procrustes model are compared to the predicted warps from the proposed model. We see that the proposed model predicts much smaller warps than the Procrustes model. 

\begin{figure}[!pt]
\centering
\setlength{\tabcolsep}{3pt}
\begin{tabular}{p{0.19\textwidth}|p{0.19\textwidth}|p{0.19\textwidth}|p{0.19\textwidth}|p{0.19\textwidth}}
\centering\scriptsize Procrustes warped template prediction & \centering\scriptsize{warped template prediction from the proposed model} & \centering\scriptsize predicted spatially correlated intensity variation & \centering\scriptsize full prediction & \centering\scriptsize observation \cr
\centering\includegraphics[width = 0.19\textwidth, trim = 60 70 30 60, clip]{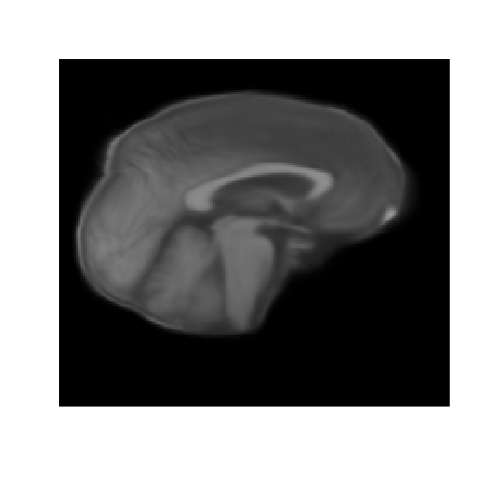} &
\centering\includegraphics[width = 0.19\textwidth, trim = 60 70 30 60, clip]{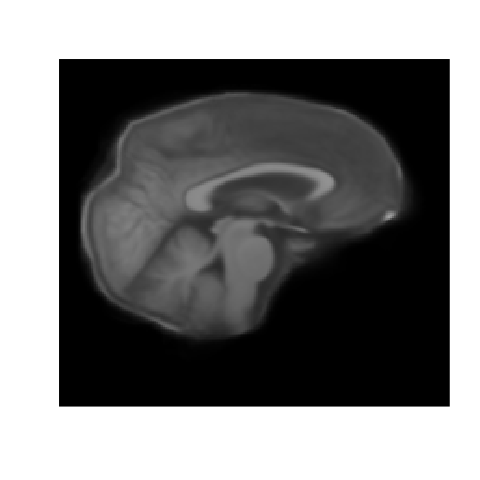} &
\centering\includegraphics[width = 0.19\textwidth, trim = 60 70 30 60, clip]{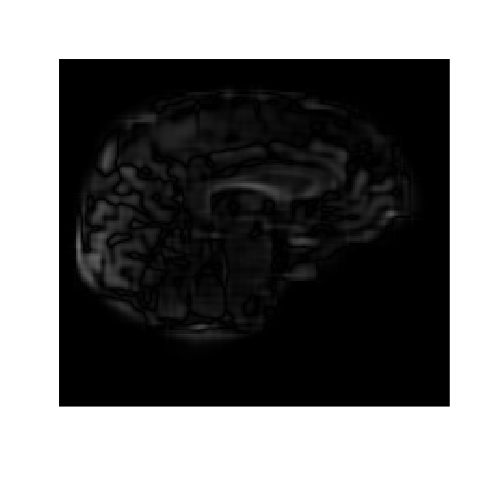} &
\centering\includegraphics[width = 0.19\textwidth, trim = 60 70 30 60, clip]{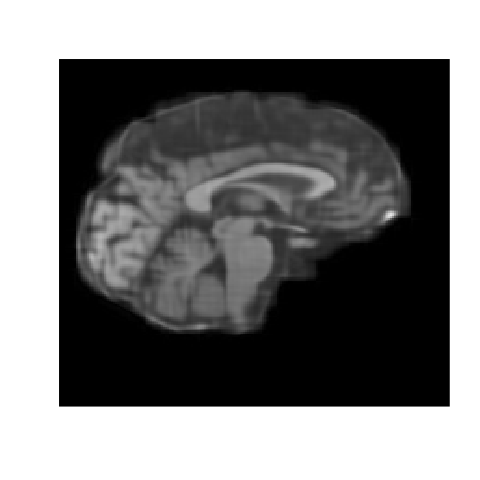} &
\centering\includegraphics[width = 0.19\textwidth, trim = 60 70 30 60, clip]{img_Brains/img_Brains2/obs1.png} \cr
\centering\includegraphics[width = 0.19\textwidth, trim = 60 70 30 60, clip]{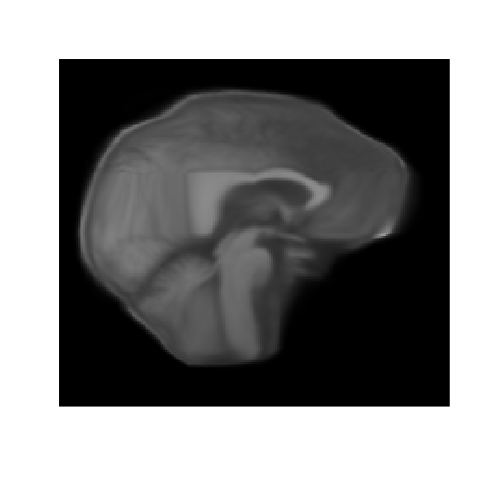} &
\centering\includegraphics[width = 0.19\textwidth, trim = 60 70 30 60, clip]{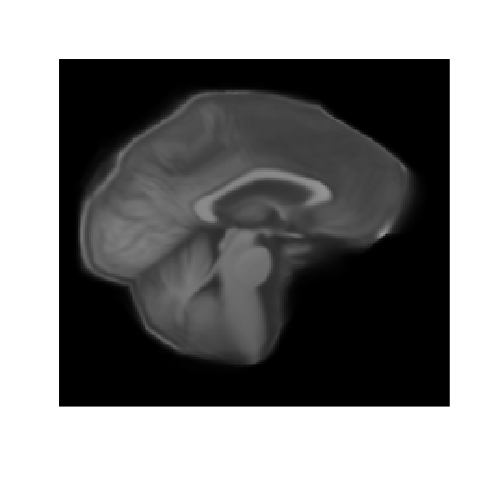} &
\centering\includegraphics[width = 0.19\textwidth, trim = 60 70 30 60, clip]{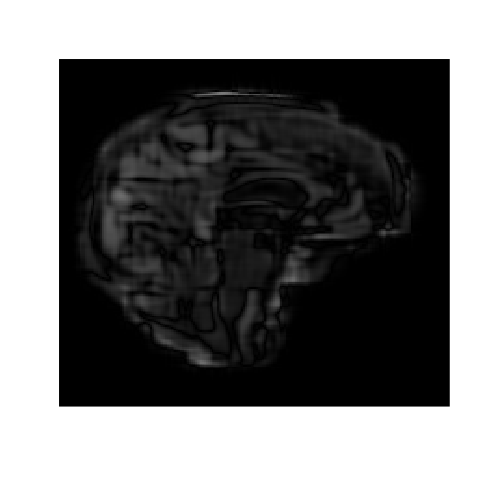} &
\centering\includegraphics[width = 0.19\textwidth, trim = 60 70 30 60, clip]{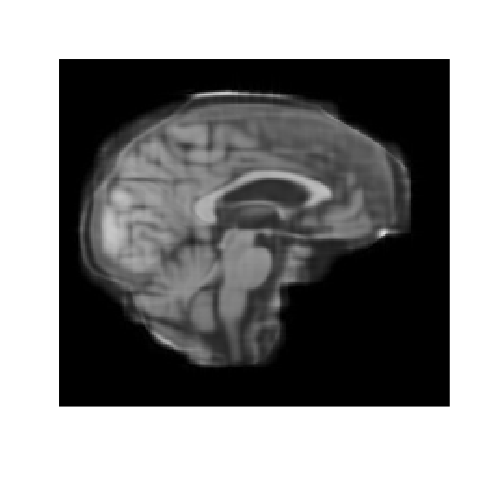} &
\centering\includegraphics[width = 0.19\textwidth, trim = 60 70 30 60, clip]{img_Brains/img_Brains2/obs2.png} \cr
\centering\includegraphics[width = 0.19\textwidth, trim = 60 70 30 60, clip]{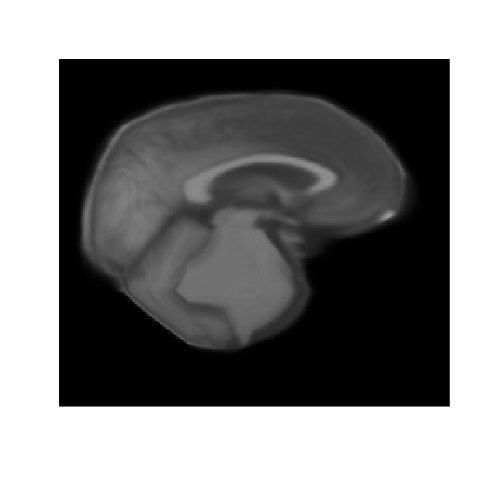} &
\centering\includegraphics[width = 0.19\textwidth, trim = 60 70 30 60, clip]{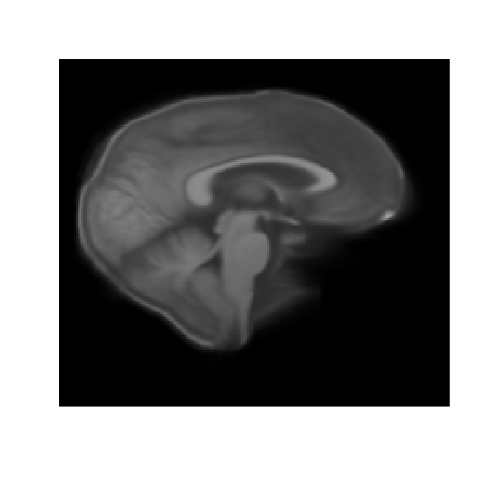} &
\centering\includegraphics[width = 0.19\textwidth, trim = 60 70 30 60, clip]{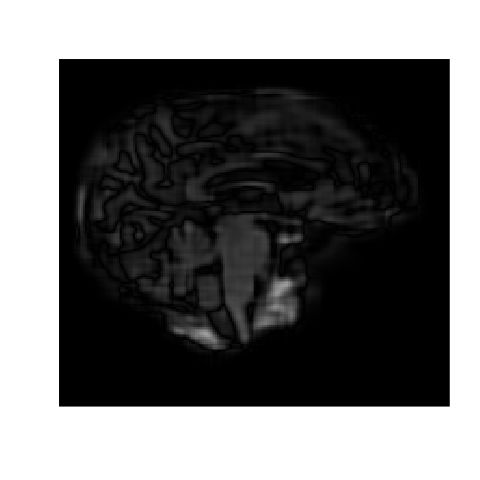} &
\centering\includegraphics[width = 0.19\textwidth, trim = 60 70 30 60, clip]{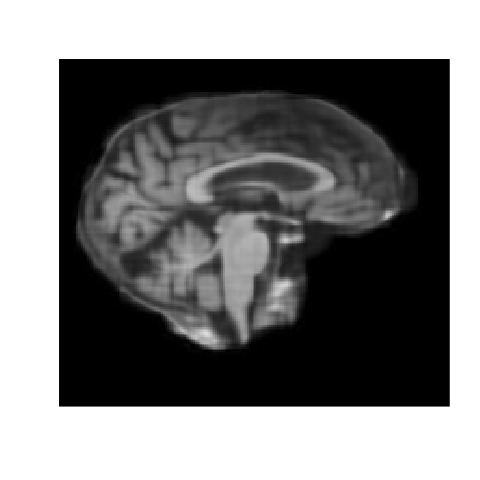} &
\centering\includegraphics[width = 0.19\textwidth, trim = 60 70 30 60, clip]{img_Brains/img_Brains2/obs3.png} \cr
\end{tabular}
\caption{Model predictions of three mid-saggital slices (rightmost column). The first two rows display the warped templates from the Procrustes model and the proposed model. The third row displays the absolute value of the predicted spatially correlated intensity variation from the proposed model. The fourth row displays the full conditional prediction given the posterior warp variables $\hat\theta(v(s, t, \hat\bw_i)) + \mathrm{E}[x_i(s, t)\,|\, \by_i, \bw_i = \hat\bw_i]$.}\label{fig:brain_pred}
\end{figure}

One of the advantages of the mixed-effects model is that we are able to predict the systematic part of the intensity variation of each image, which in turn also gives a prediction of the residual intensity variation---the variation that cannot be explained by systematic effects. In Figure~\ref{fig:brain_pred}, we have predicted the individual observed slices using the Procrustes model and the proposed model. As we also saw in Figure~\ref{fig:Dis}, the proposed model predicts less deformation of the template compared to the Procrustes model, and we see that the Brownian sheet model is able to account for the majority of the personal structure in the sulci of the brain. Moreover, the predicted intensity variation seems to model intensity differences introduced by the different MRI scanners well.

\section{Simulation study}
\label{SimStud}

\begin{figure}[!th]
\centering
\includegraphics[scale = 0.6, width = 0.30\textwidth, trim = 20 20 20 20, clip]{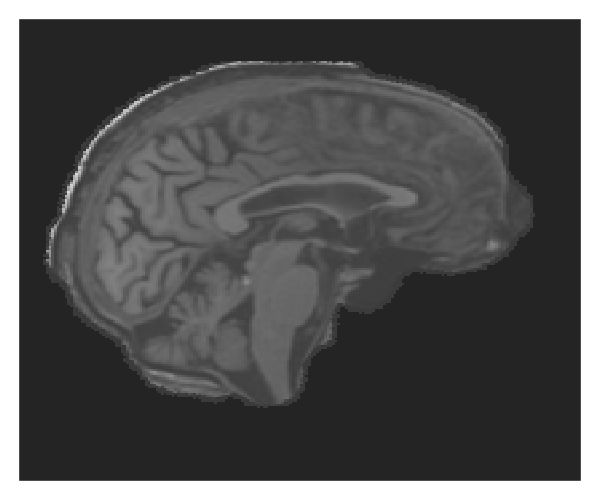}
\includegraphics[scale = 0.6, width = 0.30\textwidth, trim = 20 20 20 20, clip]{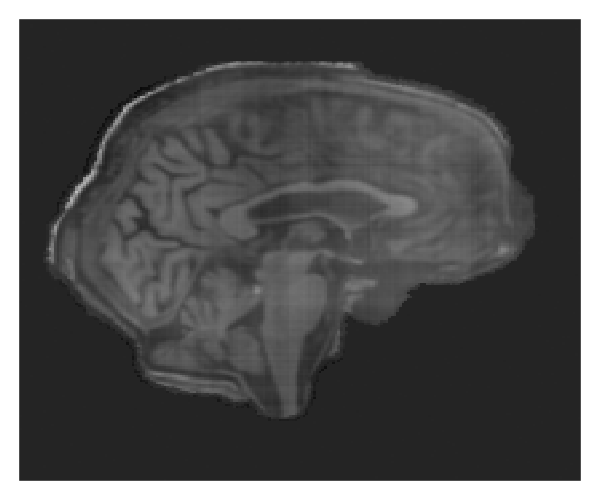}
\includegraphics[scale = 0.6, width = 0.30\textwidth, trim = 20 20 20 20, clip]{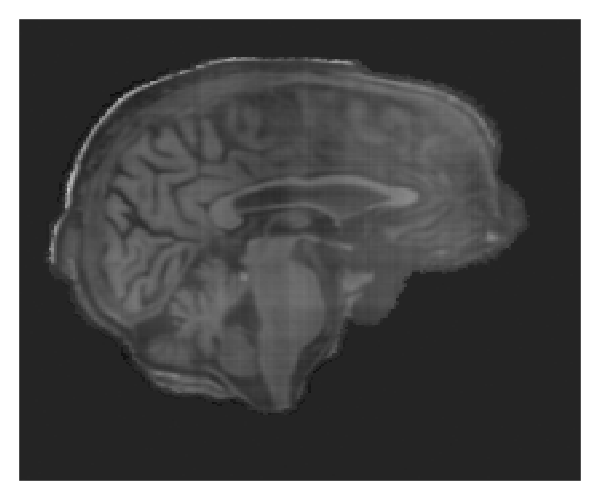}
\\[0.2em]                                                                                 
\includegraphics[scale = 0.6, width = 0.30\textwidth, trim = 20 20 20 20, clip]{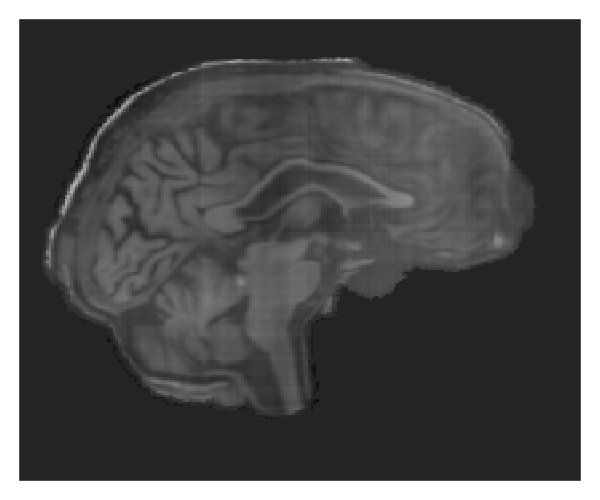}
\includegraphics[scale = 0.6, width = 0.30\textwidth, trim = 20 20 20 20, clip]{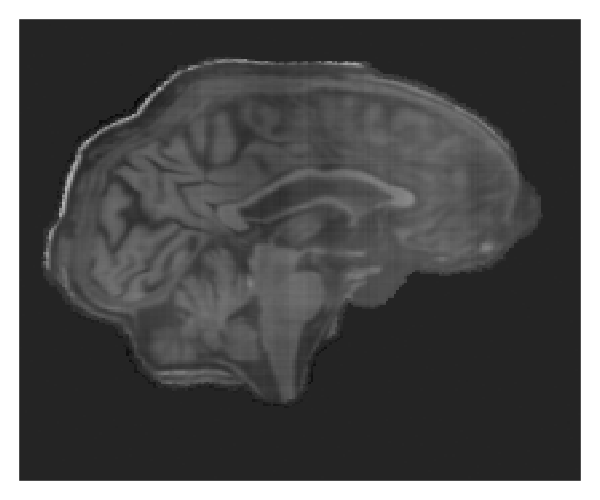}
\includegraphics[scale = 0.6, width = 0.30\textwidth, trim = 20 20 20 20, clip]{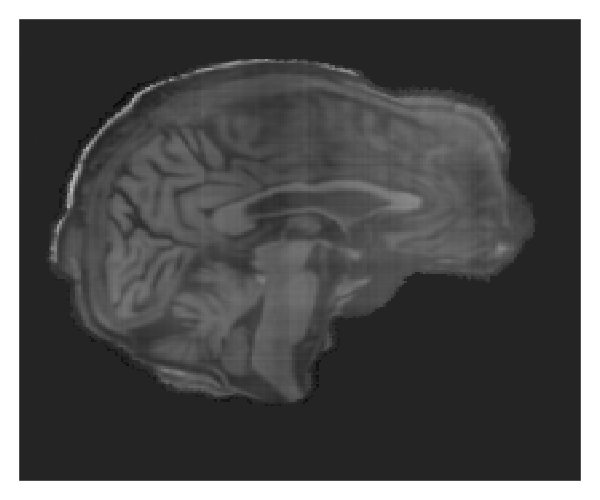}
\caption{5 examples of simulated brains. The template brain $\theta$ is shown in the upper left corner.}
\label{fig:SimEksBrains}
\end{figure}

In this section, we present a simulation study for investigating the precision of the proposed model. The results are compared to the previously introduced models: Procrustes free warp and a regularized Procrustes. 
Data are generated from model (\ref{eq:model}) in which $\theta$ is taken as one of the MRI slices considered in Section~\ref{MRI}.
The warp, intensity and the random noise effects are all drawn from the previously described multivariate normal distributions with variance parameters respectively
\[\sigma^2\gamma^2 = 0.01, \quad \sigma^2\tau^2 = 0.1, \quad \sigma^2 = 0.001\]
and applied to the chosen template image $\theta$. To consider more realistic brain simulations, the systematic part of the intensity effect was only added to the brain area of $\theta$ and not the background. As this choice makes the proposed model slightly misspecified, it will be hard to obtain precise estimates of the variance parameters. In practice, one would expect any model with a limited number of parameters to be somewhat misspecified in the presented setting. The simulations thus present a realistic setup and our main interest will be in estimating the template and predicting warp and intensity effects. Figure~\ref{fig:SimEksBrains} displays 5 examples of the simulated observations as well as the chosen $\theta$.

The study is based on 100 data sets of 100 simulated brains. For each simulated dataset we applied the proposed, Procrustes free warp and Procrustes regularized model. The regularization parameter, $\lambda$, in the regularized Procrustes model, was set to the true parameter used for generating the data $\lambda = \gamma^{-2}/2$.

\begin{figure}[!pt]
\centering
\centering\includegraphics[width = 1\textwidth]{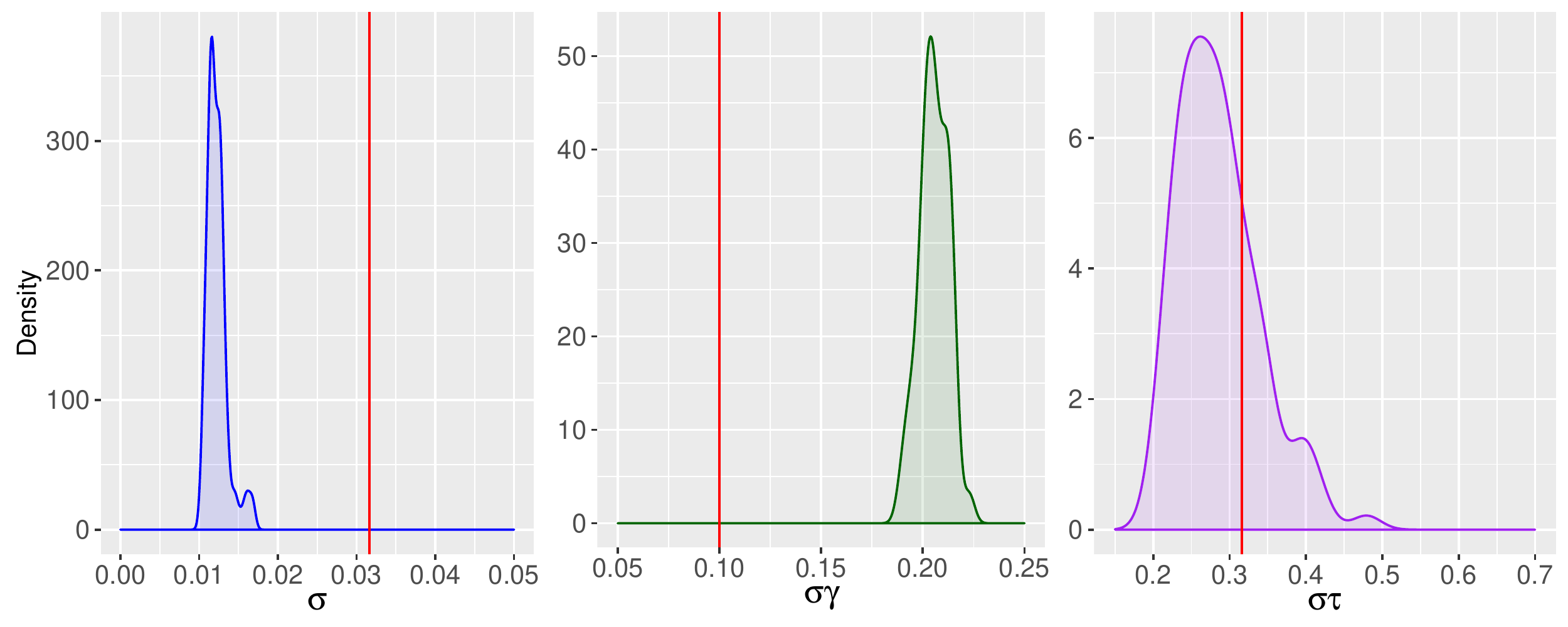}
\caption{Density plots for the estimated variance parameters in the proposed model. The red lines correspond to the true parameters.}\label{fig:VarDens}
\end{figure}

 The variance estimates based on the simulations are shown in Figure~\ref{fig:VarDens}. The true variance parameters are plotted for comparison. We see some bias in the variance parameters. While bias is to be expected, the observed bias for the noise variance $\sigma^2$ and the warp variance scale $\sigma^2\gamma^2$ are bigger than what one would expect. The reason for the underestimation of the noise variance seems to be the misspecification of the model. Since the model assumes spatially correlated noise outside of the brain area, where there is none, the likelihood assigns the majority of the variation in this area to the systematic intensity effect. The positive bias of the warp variance scale seems to be a compensating effect for the underestimated noise variance.

\begin{figure}[!pt]
\centering
\includegraphics[width = 1\textwidth]{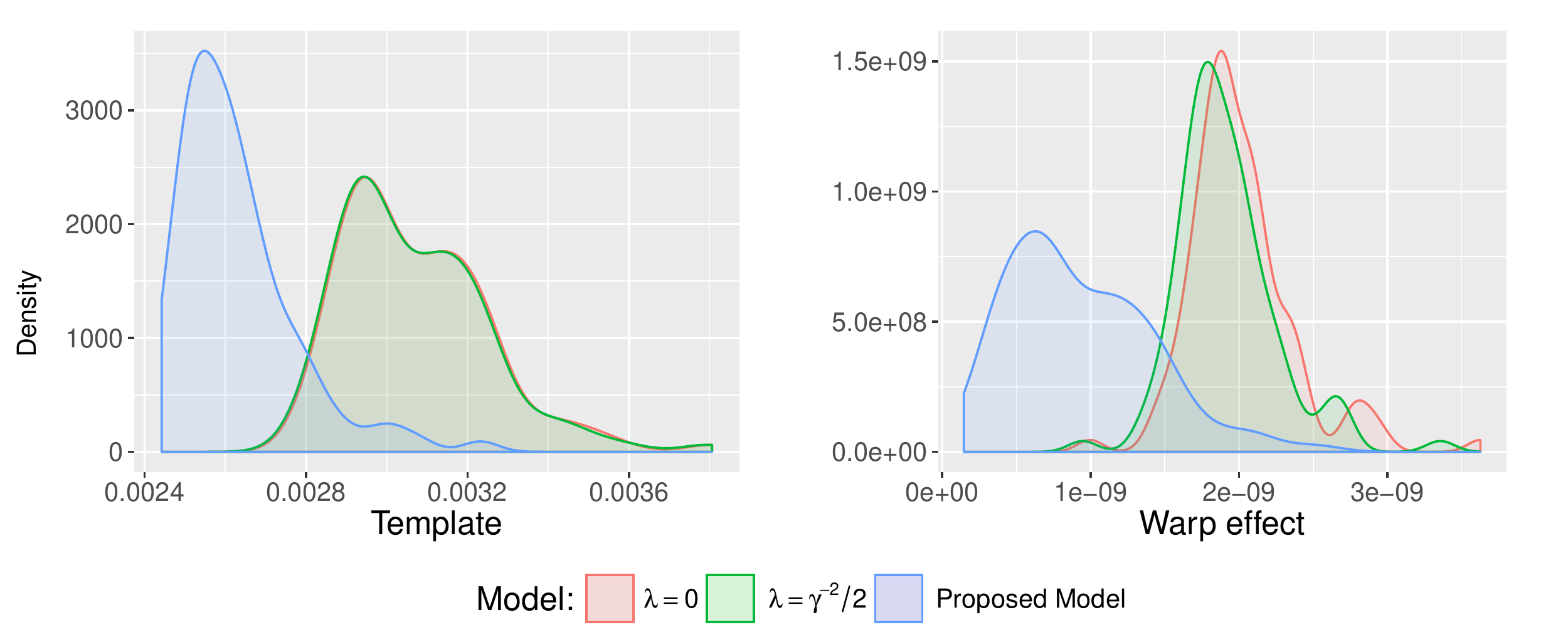}
\caption{Density plots for the mean squared differences of template and warp estimates for the three models. The plot to the left shows the density for the mean squared difference for the template effect and the plot to the right shows the mean squared difference for the warp effect. $\lambda = 0$ denotes the procrustes free warp model, $\lambda = \gamma^{-2}/2$ is the Procrustes regularized model and the blue density corresponds to the Proposed model}\label{fig:wtDens}
\end{figure}

 The left panel of Figure~\ref{fig:wtDens} shows the mean squared difference for the estimated templates $\theta$ with the three types of models. We see that the proposed model produces conisderably more accurate estimates than the alternative frameworks.

\begin{figure}[!th]
\centering
\setlength{\tabcolsep}{3pt}
\begin{tabular}{p{0.24\textwidth}p{0.24\textwidth}p{0.24\textwidth}p{0.24\textwidth}}
\centering\scriptsize True template  & \centering\scriptsize{Proposed} & \centering\scriptsize Procrustes $\lambda = 0$ & \centering\scriptsize Procrustes $\lambda = \gamma^{-2}/2$ \cr
\includegraphics[width = 0.24\textwidth, trim = 40 15 40 15, clip]{SimulationPlots/StartBrain.png} &
\includegraphics[width = 0.24\textwidth, trim = 40 15 40 15, clip]{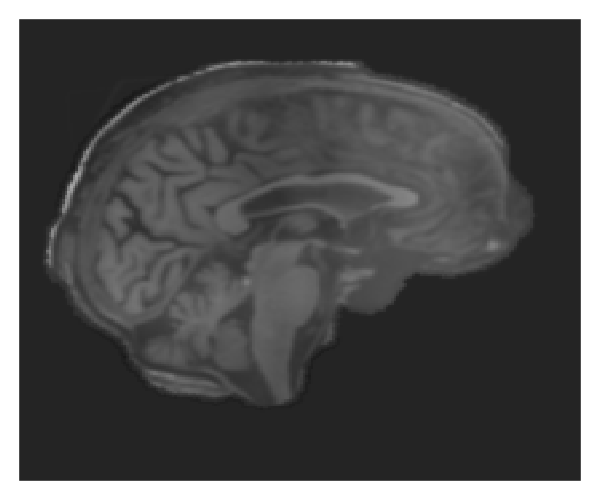} &
\includegraphics[width = 0.24\textwidth, trim = 40 15 40 15, clip]{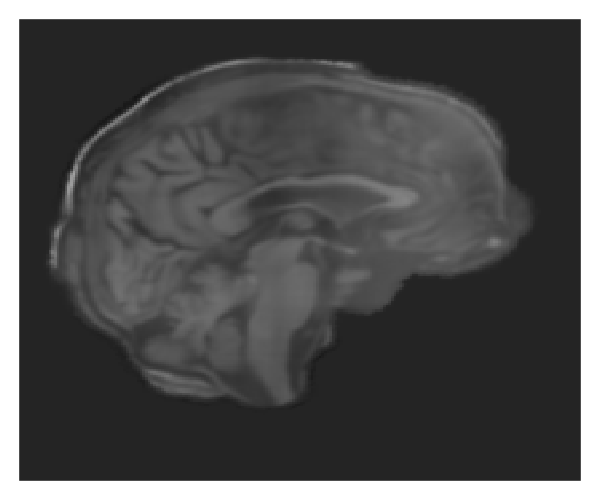} &
\includegraphics[width = 0.24\textwidth, trim = 40 15 40 15, clip]{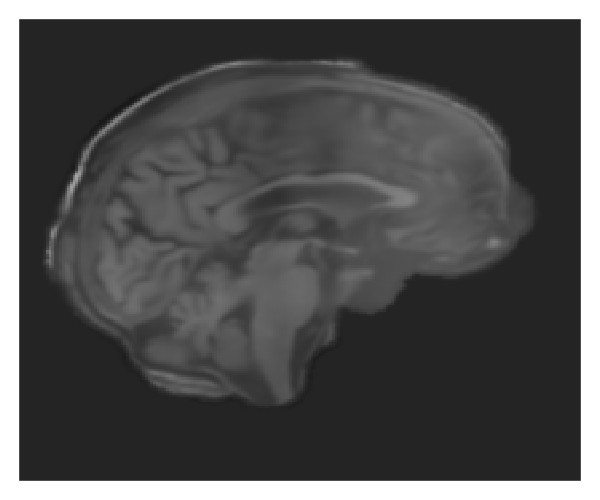}
\end{tabular}
\caption{Example of a template estimate for each of the three models. For comparison, the true $\theta$ are plotted as well.}\label{fig:tempEx}
\end{figure}

 To give an example of the difference between template estimates for the three different models, one set of template estimates for each of the models is shown in Figure~\ref{fig:tempEx}. From this example we see that the template for the proposed model is slightly more sharp than the Procrustes models and are more similar to the true $\theta$ which was also the conclusion obtained from the density of the mean squared difference for the template estimates (Figure~\ref{fig:wtDens}).

\begin{figure}[!pt]
\centering
\setlength{\tabcolsep}{3pt}
\begin{tabular}{p{0.03\textwidth}p{0.3\textwidth}p{0.3\textwidth}p{0.3\textwidth}}
\raisebox{2\normalbaselineskip}[0pt][0pt]{\rotatebox{90}{True warp effect}} &
\centering\includegraphics[width = 0.3\textwidth, trim = 40 15 40 15, clip]{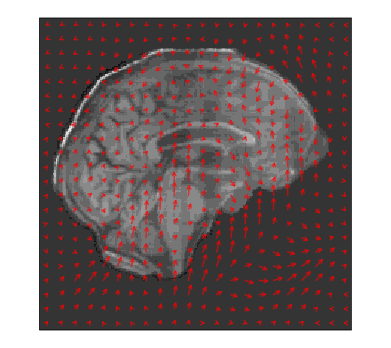} &
\centering\includegraphics[width = 0.3\textwidth, trim = 40 15 40 15, clip]{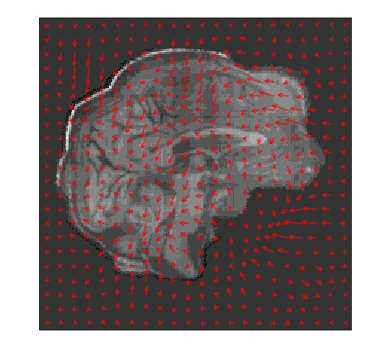} &
\centering\includegraphics[width = 0.3\textwidth, trim = 40 15 40 15, clip]{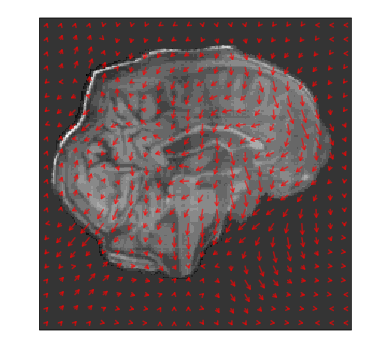} \cr
\raisebox{3.5\normalbaselineskip}[0pt][0pt]{\rotatebox{90}{Proposed}} &
\centering\includegraphics[width = 0.3\textwidth, trim = 40 15 40 15, clip]{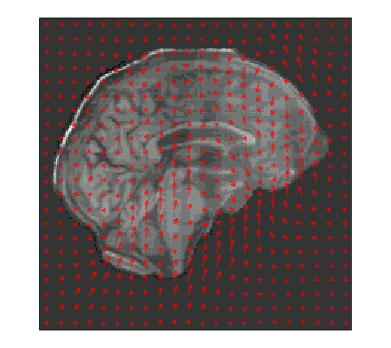} &
\centering\includegraphics[width = 0.3\textwidth, trim = 40 15 40 15, clip]{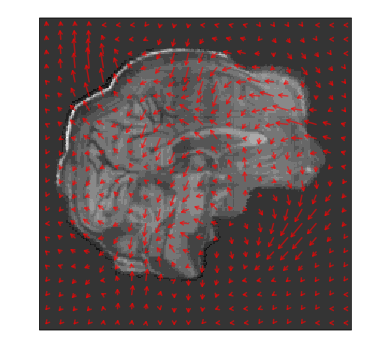} &
\centering\includegraphics[width = 0.3\textwidth, trim = 40 15 40 15, clip]{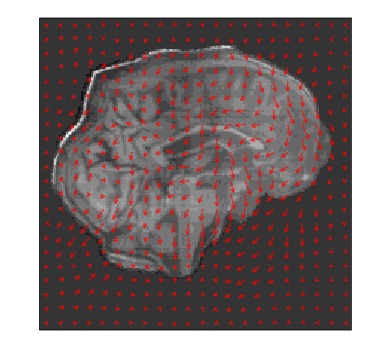} \cr
\raisebox{1\normalbaselineskip}[0pt][0pt]{\rotatebox{90}{Regularized Procrustes}} &
\centering\includegraphics[width = 0.3\textwidth, trim = 40 15 40 15, clip]{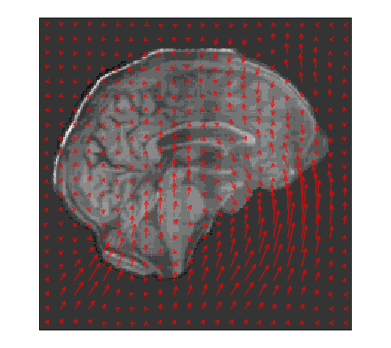} &
\centering\includegraphics[width = 0.3\textwidth, trim = 40 15 40 15, clip]{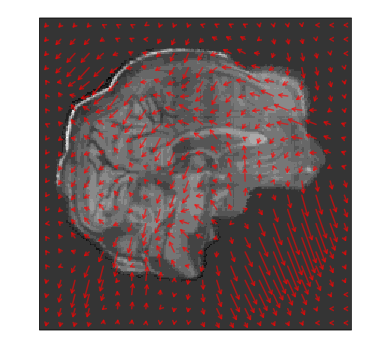} &
\centering\includegraphics[width = 0.3\textwidth, trim = 40 15 40 15, clip]{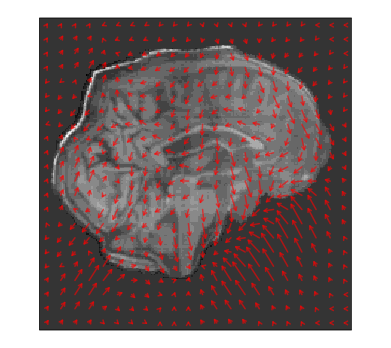} \cr 
\raisebox{1\normalbaselineskip}[0pt][0pt]{\rotatebox{90}{Procrustes free warp}} &
\centering\includegraphics[width = 0.3\textwidth, trim = 40 15 40 15, clip]{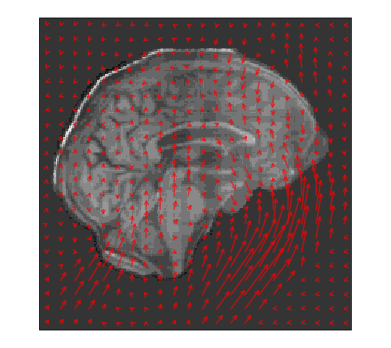} &
\centering\includegraphics[width = 0.3\textwidth, trim = 40 15 40 15, clip]{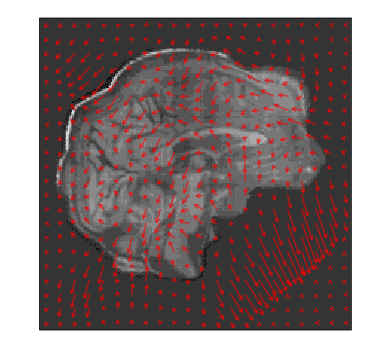} &
\centering\includegraphics[width = 0.3\textwidth, trim = 40 15 40 15, clip]{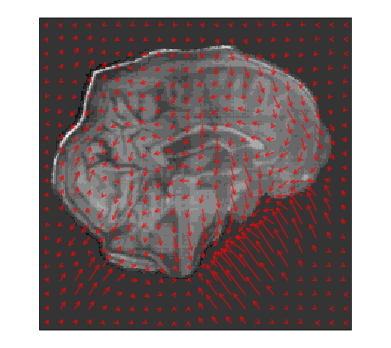}
\end{tabular}
\caption{Examples of predicted warp effect for each model. The top row shows the true warp effect, the second row the estimated warp effect of the proposed model, the third row regularized Procrustes and the final row, the Procrustes model with free warps.}\label{fig:Simpred}
\end{figure}

 The right panel of Figure~\ref{fig:wtDens} shows the mean squared prediction/estimation error of the warp effects. The error is calculated using only the warp effects in the brain area since the background is completely untextured, and any warp effect in this area will be completely determined by the prediction/estimation in the brain area. We find that the proposed model estimates warp effects that are closest to the true warps. It is worth noticing that the proposed model is considerably better at predicting the warp effects than the regularized Procrustes model. This happens despite the fact that the value for the warp regularization parameter in the model was chosen to be equal to the true parameter ($\lambda = \gamma^{-2}/2$). Examples of the true warping functions in the simulated data and the predicted/estimated effects in the different models are shown in Figure~\ref{fig:Simpred}. None of the considered models are able to make sensible predictions on the background of the brain, which is to be expected. In the brain region, the predicted warps for the proposed model seem to be very similar to the true warp effect, which we also saw in Figure~\ref{fig:wtDens} was a general tendency.

\section{Conclusion and outlook}

We generalized the likelihood based mixed-effects model for template estimation and separation of phase and intensity variation to 2D images. This type of model was originally proposed for curve data \cite{RaketSommerMarkussen}. As the model is computationally demanding for high dimensional data, we presented an approach for efficient likelihood calculations. We proposed an algorithm for doing maximum-likelihood based inference in the model and applied it to two real-life datasets.

Based on the data examples, we showed how the estimated template had desirable properties and how the model was able to simultaneously separate sources of variation in a meaningful way. This feature eliminates the bias from conventional sequential methods that process data in several independent steps, and we demonstrated how this separation resulted in well-balanced trade-offs between the regularization of warping functions and intensity variation.

We made a simulation study to investigate the precision of the template and warp effects of the proposed model and for comparison with two other models. The proposed model was compared with a Procrustes free warp model, as well as a Procrustes regularized model. Since the noise model was misspecified, the proposed methodology could not recover precise maximum likelihood estimates of the variance parameters. However, the maximum likelihood estimate for the template was seen to be a lot sharper and closer to the true template compared to alternative Procrustes models. Furthermore, we demonstrated that the proposed model was better at predicting the warping effect than the alternative models.

The main restriction of the proposed model is the computability of the likelihood function.  We resolved this by modeling intensity variation as a Gaussian Markov random field. An alternative approach would be to use the computationally efficient operator approximations of the likelihood function for image data suggested in \cite{RaketMarkussen}. This approach would, however, still require a specific choice of parametric family of covariance functions, or equivalently, a family of positive definite differential operators. An interesting and useful extension would be to allow a free low-rank spatial covariance structure and estimate it from the data. This could, for example, be done by extending the proposed model~\eqref{eq:model} to a factor analysis model where both the mean function and intensity variation is modeled in a common functional basis, and requiring a specific rank of the covariance of the intensity effect. Such a model could be fitted by means of an EM algorithm similar to the one for the reduced-rank model for computing functional principal component analysis proposed in \cite{james2000principal}, and it would allow simulation of realistic observations by sampling from the model.

For the computation of the likelihood function of the nonlinear model, we relied on local linearization which is a simple well-proven and effective approach. In recent years, alternative frameworks for doing maximum likelihood estimation in nonlinear mixed-effects models have emerged, see \cite{carpenter2016stan} and references therein. An interesting path for future work would be to formulate the proposed model in such a framework that promises better accuracy than the local linear approximation. This would allow one to investigate how much the linear approximation of the likelihood affects the estimated parameters. In this respect, it would also be interesting to compare the computing time across different methods to identify a suitable tradeoff between accuracy and computing time.

The proposed model introduced in this paper is a tool for analyzing 2D images. The model, as it is, could be used for higher dimensional images as well, but the analysis would be computationally infeasible with the current implementation. To extend the proposed model to 3D images there is a need to devise new computational methods for improving the calculation of the likelihood function.


\bibliographystyle{abbrv}

\end{document}